\documentclass[]{elsarticle}
\usepackage{lineno,hyperref}
\modulolinenumbers[5]
\usepackage{amsmath}
\usepackage{graphicx}
\def\argmax{\mathop{\rm arg\,max}\limits}
\begin{document}
\begin{frontmatter}
\title{Recovering the number of clusters in data sets with noise features using feature rescaling factors\footnote{This is an accepted manuscript and now published in Information Sciences 324 (2015), 126-145, doi:10.1016/j.ins.2015.06.039.\\
\textcopyright 2015.  This manuscript version is made available under the CC-BY-NC-ND 4.0 license http://creativecommons.org/licenses/by-nc-nd/4.0/}}
\author[mymainaddress]{Renato Cordeiro de Amorim\corref{mycorrespondingauthor}}
\cortext[mycorrespondingauthor]{Corresponding author at School of Computer Science, University of Hertfordshire, College Lane Campus, Hatfield AL10 9AB, UK. Phone:+44 01707 286160 Fax:+44 01707 284115.}
\ead{r.amorim@herts.ac.uk}
\author[mysecondaryaddress]{Christian Hennig\corref{esprc}}
\address[mymainaddress]{School of Computer Science, University of Hertfordshire, College Lane Campus, Hatfield AL10 9AB, UK.}
\address[mysecondaryaddress]{Department of Statistical Science, University College London, Torrington Place, London WC1E 6BT, UK.}
\cortext[esprc]{The work of this author was supported by EPSRC grant EP/K033972/1}
\ead{c.hennig@ucl.ac.uk}

\begin{abstract}
In this paper we introduce three methods for re-scaling data sets aiming at improving the likelihood  of clustering validity indexes to return the true number of spherical Gaussian clusters with additional noise features. Our method obtains feature re-scaling factors taking into account the structure of a given data set and the intuitive idea that different features may have different degrees of relevance at different clusters.

We experiment with the Silhouette (using squared Euclidean, Manhattan, and the p$^{th}$ power of the Minkowski distance), Dunn's, Calinski-Harabasz and Hartigan indexes on data sets with spherical Gaussian clusters with and without noise features. We conclude that our methods indeed increase the chances of estimating the true number of clusters in a data set.
\end{abstract}
\end{frontmatter}
\section{Introduction}

Clustering is one of the most popular tasks in data analysis. It aims to reveal a class structure in a data set by partitioning it in an unsupervised manner. 

In this paper we address the fundamental issue of estimating the number of clusters $K$ in a data set. This particular problem has raised considerable research interest over the years, but it is not without controversies. It is a very active field of research \citep{chiang2010intelligent,arbelaitz2012extensive,jain1988algorithms,steinley2006k}, but due to the lack of a generally accepted definition of what a ``cluster'' is there are no unified standards against which it can be assessed.

A cluster is a homogeneous group of entities. While entities in the same cluster are supposed to be homogeneous, according to some notion of similarity, entities in different clusters are expected to be heterogeneous. This is a rather loose definition for the term cluster, which does not help much defining the true number of clusters for a given data set. In order to make the problem more precise, in the present paper we are interested in finding clusters in the sense of K-Means but with additional non-informative (``noise'') features, that is, clusters that are on the cluster-defining features approximately spherical and compact, with similar within-cluster variation, and that can be approximated well by Gaussian distributions. This clustering problem can be solved more precisely and in a more meaningful way if there is some degree of separation between clusters (i.e., distance between the cluster's high density areas), because otherwise, even if the number of clusters is correctly diagnosed, strong overlap between Gaussian distributions means that points cannot be reliably assigned to the Gaussian component that generated them. We agree with \citep{Hennig_Liao_2013} that different clustering methods are appropriate for different clustering aims, and that when carrying out a cluster analysis, researchers need to define more precisely what kind of clusters they are interested in. In line with this thought, the above somewhat restrictictive cluster definition can help researchers to decide whether the methods presented here are suitable, instead of claiming that we could solve the general problem of clustering and estimating the numbers of clusters.  

This defines what we mean by ``true'' clusters in the following, acknowledging that it does not yield a general definition of the clustering problem, but rather a working definition for one of many possible ways to understand the term ``cluster'', at which the methods discussed here are aimed. The methods that we propose actually allow for more general than spherical cluster shapes as long as feature re-scaling transforms the cluster shapes into (approximately) spherical ones. 

Various clustering algorithms, some explained in Sections \ref{Sec:Background} and \ref{Sec:FeatureWeighting}, are unable to determine the number of clusters in a given data set, and in fact request this number to be specified beforehand. In scenarios in which this number is not known, a popular solution is to run a given clustering algorithm using different values for the number of clusters and then analyse the generated clusterings afterwards. The process of estimating how well a partition fits the structure underlying the data is often called ``cluster validation'' \citep{arbelaitz2012extensive,halkidi2001clustering}. After all feasible possibilities are analysed the number of clusters that generated the best partition, according to a clustering validation index, is selected.

Note that it cannot be taken for granted that the problem of finding the true number of clusters coincides with finding the clustering solution that produces the best clustering in terms of the misclassification rate or the adjusted Rand index \citep{hubertarabie1985rand}. For example, it may be that if there are two true clusters, the clustering method splits the data set up incorrectly if indeed $K=2$ is used as the number of clusters, whereas for $K=3$ one cluster coincides perfectly with one true cluster and the other true cluster is split into two found clusters, which for many applications and in terms of the adjusted Rand index may be seen as the better solution. In this paper we aim to address both views, finding the true $K$, and finding the best clustering.
 
The quantity of noise features in a data set is an important concern. It is not uncommon to have data sets containing entities characterized by features, with some of the latter being irrelevant to the problem at hand. Generally speaking, noise features, together with the degree of overlap between clusters, are the factors with the greatest impact on clustering validation indexes performance \citep{arbelaitz2012extensive,chiang2010intelligent}, with a small inclusion of 10\% noise features having already a considerable impact on such indexes \citep{arbelaitz2012extensive}.
 
In our experiments we simulate irrelevant features by adding features generated from uniform random values to our data sets. Difficulties in the estimation of the number of clusters raised by the presence of noise features in a data set have been considered before \citep{dudoit2002prediction}, however, there is still a view that the issue raised by noise features deserves more consideration \citep{chiang2010intelligent}. 

The main contribution of this paper is to present three methods to re-scale data sets in such a way that cluster validity indexes become more likely to return the true number of clusters. Our experiments focus on versions of the most popular partitioning algorithm, K-Means, and the comparison of the performances of each index before and after re-scaling. 

Section \ref{Sec:Background} reviews K-Means and a number of validation indexes. The methodological core of the paper is Section \ref{Sec:FeatureWeighting}, in which we introduce a version of K-Means incorporating feature weighting and more general Minkowski metrics \citep{cordeiro2011minkowski}. Different versions of feature re-scaling with and without re-clustering at the end are proposed for use with the validation indexes. In Section \ref{Sexp} we present our simulation study and sicuss its results, followed by a conclusion.
\section{Background and related work}
\label{Sec:Background}
\subsection{K-Means}\label{Skmeans}
K-Means \citep{macqueen1967some,ball1967clustering} is arguably the most popular partitioning clustering algorithm \citep{jain2010data,steinley2006k}. Given a data set $Y$ of $V$-dimensional entities $y_i \in Y$, for $i=1, 2, ..., N$, K-Means generates $K$ non-empty disjoint clusters $S=\{S_1, S_2, ..., S_K\}$ around the centroids $C=\{c_1, c_2, ..., c_K\}$, by iteratively minimising the sum 
\begin{equation}
\label{Eq:K_Means}
W_K=W(S,C) = \sum_{k=1}^{K}\sum_{i \in S_{k}} d(y_i, c_{k})
\end{equation}
of the within-cluster distance between entities and centroids. Each centroid $c_k$ uniquely represents a cluster $S_k$ and is sometimes called its prototype. The K-Means criterion above returns an index representing how good a clustering is, the lower the better.
$d(y_i,c_k)$ in (\ref{Eq:K_Means}) represents the distance between entity $y_i$ and the centroid $c_k$. In the original K-Means, this distance measure is the squared Euclidean distance given by $d(y_i, c_k)=\sum_{v \in V} (y_{iv} - c_{kv})^2$, minimising the square error criterion. Other distance measures are possible, such as the Manhattan distance given by $d(y_i, c_k)=\sum_{v \in V} |y_{iv} - c_{kv}|$, although only with the squared Euclidean distance the cluster centroids minimising $W$ are actually the within-cluster means. In the present paper, we will not only consider the Euclidean distance, but also the Manhattan distance and the p$^{th}$ power of the Minkowski distance $d_p(y_i,c_k) = \sum_v |y_{iv}-c_{kv}|^p$ for various values of $p$, because with a suitable choice of $p$ this has been found to work well with noise features, see \citep{cordeiro2011minkowski} and Section
\ref{Sec:FeatureWeighting}.

The p$^{th}$ power of the Minkowski distance is chosen here by analogy to the use of the squared Euclidean distance, rather than the Euclidean distance itself, in the original K-Means.

The minimisation of (\ref{Eq:K_Means}) has three simple steps, iterated until convergence.
\begin{enumerate}
\itemsep0em 
\item Select the values of $K$ entities $y_i \in Y$ as initial centroids $c_1, c_2, ..., c_K$. The initial entities may be chosen at random, but better strategies are available, see Section \ref{Sintel}. Set $S=\{\emptyset \}$.
\item Assign each entity $y_i \in Y$ to the cluster $S_k$ represented by $c_k$, the closest centroid to $y_i$.
\item If there are no changes to $S$, stop and output $S$ and $C$. Otherwise update each centroid $c_k$ to the centre of its cluster $S_k$ and go to Step 2.
\end{enumerate}
Due to its popularity, K-Means has been applied in a wide range of fields including bioinformatics, data mining, natural language processing \citep{vedaldi2010vlfeat,steinley2006k,jain2010data,sturn2002genesis,huang2008weighting,gasch2002exploring,mirkin2012clustering}. The popularity of K-Means has also exposed some weaknesses. Due to its greedy nature, K-Means does not guarantee a convergence to a global optimum, with its final clustering heavily dependent on the initial centroids. K-Means treats each feature equally, regardless of its actual degree of relevance to the problem, and requires the number of clusters in the data set, $K$, to be known beforehand.
%

There are a number of cluster validity indexes (CVIs) that could be used to estimate $K$. It would be impossible for us to experiment with all of them, so we have chosen some indexes that can be considered as most popular and that work with general distance measures, which will be introduced below.

%
\subsection{Silhouette index}
\label{Subsec:Silhouette}
The Silhouette width \citep{rousseeuw1987silhouettes} is a ratio-type index that is based on silhouette values for every entity $y_i$ measuring how well $y_i$ fits into the cluster to which it is assigned, by comparing the within-cluster cohesion, based on the distance to all entities in the same cluster, to the cluster separation:
\begin{equation}
s(y_i) = \frac{b(y_i)-a(y_i)}{max\{a(y_i), b(y_i)\}},
\end{equation}
where $a(y_i)$ is the average dissimilarity of $y_i \in S_k$ to all other $y_j \in S_k$, $b(y_i)$ the minimum dissimilarity over all clusters $S_l$, to which $y_i$ is not assigned, of the average dissimilarities to $y_j \in S_l, l\not=k$. Therefore, $-1\leq s(y_i) \leq 1$. If $s(y_i)$ is around zero, the entity $y_i$ could be assigned to another cluster without making cluster cohesion or separation any worse. A negative $s(y_i)$ suggests that $y_i$'s cluster assignment is damaging to cluster cohesion and separation, whereas an $s(y_i)$ closer to 1 means the opposite. We can then quantify the validity of the whole clustering by the Silhouette index, defined as $\frac{1}{N}\sum_{i \in Y} s(y_i)$. 

The literature indicates that there is no sole cluster validity index with a clear advantage over the others in every case \citep{bezdek1998some}. However, the Silhouette width index has performed well in many comparative experiments \citep{arbelaitz2012extensive,pollard2002method}. 

The Silhouette index can work with any distance measure. In the present work, the Silhouette index was applied using general Minkowski distances including the Euclidean and Manhattan distances, in line with using general Minkowski-distances in our method.

\subsection{Dunn's index}
Dunn's index \citep{dunn1973fuzzy} $D$ is defined as the ratio of the smallest distance between clusters, which estimates the separation of clusters, and the maximum cluster diameter, which estimates its cohesion. This index allows for general distance measures and was applied here with the Euclidean as well as general Minkowski distances.

Dunn's index is not without flaws. Possibly the most relevant in relation to our experiment is its sensitivity to the information in noise features. However, this index does provide a rich and very general structure for defining cluster validity indexes for different types of clusters \citep{bezdek1998some}.
\subsection{Calinski-Harabasz index}
The Calinski-Harabasz index \citep{calinski1974dendrite} is a popular index using a ratio of a between-cluster-means and a within-cluster sum of squares statistic:
\begin{equation}
CH=\frac{(T-W_K)/(K-1)}{W_K/(N-K)},
\end{equation}
where $T$ is the data scatter defined by $T=\sum_{i=1}^N \sum_{v \in V} (y_{iv}-\bar{y_v})^2$, $W_K$ defined as in \eqref{Eq:K_Means}. Being a winner in the classical comparative study in \citep{milligan1985examination} made this index popular. Due to its original motivation by the statistical theory behind F-test comparing groups in multivariate ANOVA, this index uses the Euclidean distance exclusively.
\subsection{Hartigan index}
The Hartigan index \citep{hartigan1975clustering} is a heuristic rule of thumb with a considerable amount of success. This index has been found in one study to be the best performer for finding the number of clusters \citep{chiang2010intelligent}. As $CH$, the Hartigan-index is also based on the Euclidean within-cluster sum of squares, and considers its change when increasing $K$:


\begin{equation}
HK=(W_K/W_{K+1}-1)(N-K-1).
\end{equation}
In the original paper, the lowest $K$ to yield $HK \leq 10$ was proposed as the optimal choice \citep{hartigan1975clustering}. We follow this approach in our experiments, however, when we are unable to find a $K$ that meets this criteria, we simply choose the $K$ whose difference in $HK$ for $K$ and $K+1$ is the smallest.



In practice, all these indexes need a maximum value of $K$ to be examined, which we chose as 20 in our experiments.
\subsection{Intelligent K-Means}\label{Sintel}
Intelligent K-Means (iK-Means) \citep{mirkin2012clustering} addresses the intrinsically related issues of finding the number of clusters, $K$, as well as good initial centroids $c_1, c_2, ..., c_K$ for K-Means. It does so by extracting anomalous patterns from a given data set, one at a time, using the algorithm below. This needs a tuning constant $\theta$, an integer value with the interpretation of the minimum accepted initial cluster size.
\begin{enumerate}
\itemsep0em 
\item Set $c_c=\bar y$, the centre of data set $Y$.
\item Set $c_t=\argmax_{y_i} d(y_i,c_c)$, the farthest entity $y_i \in Y$ from $c_c$.
\item Run K-Means supplying $c_c$ and $c_t$ as initial centroids, forming clusters $S_c$ and $S_t$. The aim is to find a cluster $S_t$ anomalous in relation to $c_c$, hence the latter should not be allowed to move during the K-Means iterations, i.e. in step 3 of the algorithm given in Section \ref{Skmeans}, $c_c$ is kept fixed.
\item If $|S_t|\geq \theta$, $|S_t|$ denoting the number of elements of $S_t$, add $c_t$ to $C_{init}$. In any case remove the entities $y_i \in S_t$ from $Y$.
\item If there are still entities to cluster, go to Step 2.
\item Run K-Means using the centroids in $C_{init}$ as initial centroids and set $K$ to $|C_{init}|$.
\end{enumerate}
The iK-Means algorithm has been compared favourably to a number of other algorithms \citep{chiang2010intelligent,de2013empirical,chiang2007experiments}. However, a variant of it, with a pre-specified $K$ and no removal of singletons, was compared by Steinley et al. \citep{steinley2007initializing}. These modifications lead the algorithm to reach poor results. 

In its original version, iK-Means has a good performance recovering clusters and centroids, but the authors acknowledge that it may drastically overestimate the number of clusters \citep{chiang2010intelligent}. Here we use iK-Means' choice of $K$ not as a recommended estimator, but rather as a tool to reduce the domain used to search for $K$. In our baseline experiments (see Section \ref{Sexp}),
we produce results for other indexes searching for $K$ in the interval $[2,20]$. In the experiments with our feature re-scaling method we produce experiments searching for $K$ between 2, and the lowest between 20 and the number of clusters found by iK-Means when $\theta$ is set to zero.

We discuss this further in Section \ref{Sec:FeatureWeighting} when presenting the intelligent Minkowski Weighted K-Means.
\section{K-Means with feature weighting and Minkowski distance}
\label{Sec:FeatureWeighting}
\subsection{Minkowski weighted K-Means}
\label{Sec:MWK}
The Minkowski weighted K-Means (MWK-Means) \citep{cordeiro2011minkowski} was designed to extend the original K-Means algorithm by addressing one of its main weaknesses, namely dealing with data sets containing many irrelevant features. In the original K-Means each of the features in $V$ is treated equally, meaning that an irrelevant feature would have the same contribution to the clustering as a relevant feature. Even among relevant features there may be very different degrees of relevance and this should be taken into account by the algorithm. Note that we take a feature as ``relevant'' here if it provides information about the true clustering as defined above; in some applications though, the relevance of features may depend on their meaning and the aim of clustering \citep{Hennig_Liao_2013}.

The Minkowski metric, defined as $d_p(y_{i}, c_{k}) = \sqrt[p]{\sum_{v \in V} |y_{iv} - c_{kv}|^{p}}$ for the $V$-dimensional $y_i$ and $c_k$, is a generalisation of the Manhattan ($p=1$), Euclidean ($p=2$) and Chebyshev ($p\rightarrow \infty$) metrics. The MWK-Means actually uses the $p^{th}$ power of the Minkowski metric, not involving the root, in analogy to the use of the squared Euclidean distance by standard K-Means.

The use of weights is directly related to the concept of the distance measure in use. We have generalized the $p^{th}$ root of the Minkowski metric by introducing a weight $w$, also to the power of $p$:
\begin{equation}
\label{Eq:AdjustedDistanceMeasure}
	d_p(y_{i}, c_{k}) = \sum_{v \in V} w^{p}_{kv} |y_{iv} - c_{kv}|^{p}.
\end{equation} 
The use of the Minkowski metric brings two benefits to MWK-Means. It allows for the recovery of clusters formed in shapes other than spherical, and it allows us to interpret the weights as feature scaling factors for any $p$. The latter is not true for other feature weighting methods for K-Means based on a powered weight, such as Weighted K-Means and the attributes-weighting clustering algorithm \citep{huang2005automated,chan2004optimization}.

By using the distance (\ref{Eq:AdjustedDistanceMeasure}) in the K-Means criterion (\ref{Eq:K_Means}), we obtain the MWK-Means criterion:
\begin{equation}
\label{Eq:MWK-Means}
W_{p}(S,C,w)=\sum_{k=1}^{K}\sum_{i \in S_k}\sum_{v=1}^{V}w^{p}_{kv}|y_{iv}-c_{kv}|^{p},
\end{equation}
where $w_{kv}$ represents the weight of a particular feature $v$ for cluster $S_k$, and is subject to $\sum_{v \in V} w_{kv} =1$. Such feature weights are obtained following the intuitive assumption that the less dispersion a given feature has within the clusters, the higher its weight should be. Note that the K-Means criterion can be related to an implicit assumption that clusters are sperical, i.e., all variables have about the same variance within clusters \citep{jain2010data}, which means that it appears appropriate that variables that vary strongly within ``true'' clusters are weighted down.
We calculate the feature dispersion of variable $v$ within cluster $k$ by $D_{kv}=\sum_{i \in S_k}|y_{iv} - c_{kv}|^p$, and the weights, introduced in
 \citep{cordeiro2011minkowski}:
\begin{equation}
\label{Eq:FeatureWeights}
w_{kv}=\frac{1}{\sum_{u\in V}[{D_{kv}}/{D_{ku}}]^{1/(p-1)}},
\end{equation}
see Section \ref{Sec:RescalingMethod} for a derivation.
To avoid issues related to divisions by zero in (\ref{Eq:FeatureWeights}), as well as a weight $w_{kv}$ of zero for a feature whose dispersion is zero in a particular cluster $S_k$ (which would be a rather informative feature), we always add a constant to each dispersion $D_{kv}$, namely the average dispersion over all features.

The iterative minimisation of (\ref{Eq:MWK-Means}) is very similar to the minimisation of the K-Means criterion itself (\ref{Eq:K_Means}) and has a single extra step, to calculate the feature weights.
\begin{enumerate}
\label{Alg:MWK}
\item Select the value of $K$ entities $y_i \in Y$ as initial centroids $c_1, c_2, ..., c_K$. Set each weight $w_{kv}=1/|V|$.
\item Assign each entity $y_i \in Y$ to the cluster $S_k$ represented by the closest centroid $c_k$ as per (\ref{Eq:AdjustedDistanceMeasure}).
\item Update each centroid $c_k$ for $k=1, 2, ..., K$ to the Minkowski centre of its cluster $S_k$, see below. If there are no changes, stop and output $S$, $C$ and $w$.
\item Update all feature weights $w_{kv}$ applying Equation (\ref{Eq:FeatureWeights}); go back to Step 2.
\end{enumerate}
MWK-Means requires the calculation of centroids representing each cluster, minimising the sum of the $p$th power of the Minkowski-distances to the centroids,
called ``Minkowski centres'' here. This can be computed separately for each feature. To compute them is quite straightforward for $p=1$ (leading to the componentwise median) and $p=2$ (leading to the mean), but less so for other values of $p$. In our experiments we have constrained $p\geq 1$. In this case $\gamma_v(\mu) =\sum_{i \in S_k} |y_{iv} - \mu|^p$ is a U-shaped curve with a minimum in the interval $[\min_i(y_{iv}), \max_i(y_{iv})]$ \citep{cordeiro2011minkowski}. The minimum $\mu_{kv}$ can be found by standard methods for convex optimisation; a very straightforward approach, which we used here, starts from the variable mean and improves it stepwise by moving it by some fixed amount (0.001, say) per step to the side in which $\gamma_v$ is reduced. 

MWK-Means converges in a finite number of iterations because at each one of them (\ref{Eq:MWK-Means}) decreases, whereas the number of different partitions is finite \cite{cordeiro2011minkowski}. Very much like K-Means this is a non-deterministic algorithm depending heavily on the initial centroids. See Section \ref{SreiMWK} for the use of the iK-Means principle for initialization.

\subsection{Details on feature re-scaling}
\label{Sec:RescalingMethod}
The re-scaling of features in a data set is common practice during the data pre-processing step of virtually any machine learning algorithm. The aim is to make sure that features whose values originally specified using larger numbers do not dominate features that may be equally, or indeed even more relevant, but that have been originally specified using a smaller scale. In all of our experiments we started by standardizing each of our data sets as follows, in order to make the scales comparable: 
\begin{equation}
\label{Eq:Standardize}
y_{iv}=\frac{y_{iv} - \bar{y_v}}{max(y_v)-min(y_v)},\ i=1,\ldots,N,\ v \in V, 
\end{equation}
where $\bar{y_v}$ represents the average of feature $v$ over each entity $y_i \in Y$. Our choice of standardizing by the range, rather than the most popular use of the standard deviation (the \textit{z}-score), was because the latter favours unimodal distributions. For example, given two features, $n$ (unimodal) and $m$ (bimodal) with the same range, the latter would normally have a higher standard deviation than the former. This property would lead to smaller values in $m$ than in $n$ after standardization, even though clustering would normally target groupings associated to the modes present in $m$.

However, we are not convinced that having each feature treated equally as a consequence of applying (\ref{Eq:Standardize}) is the most desirable outcome. We would prefer the scale of a given feature $v$ to be related to its actual degree of relevance to the problem at hand. In fact, going even further, we would prefer this re-scaling process to take into account the structure of $Y$ so that this re-scaling process takes into account the different degrees of relevance a feature $v$ may have at different clusters $S_k \in S$.

The above surely sounds rather intuitive. However, applying different re-scaling at different clusters may be problematic, because it may affect distances between different clusters. 



The algorithm minimising the MWK-Means criterion (\ref{Eq:MWK-Means}) described in Section \ref{Sec:FeatureWeighting} applies partial optimisation for $S$, $C$, and $w$. Here we show that (\ref{Eq:FeatureWeights}) does indeed minimise (\ref{Eq:MWK-Means}) for a fixed $S$ and $C$, which was not clearly shown in \citep{cordeiro2011minkowski}.

The minimisation of (\ref{Eq:MWK-Means}) is subject to $\sum_{v \in V} w_{kv}=1$ for $k=1, 2, ..., K$, and a crisp clustering where any given entity $y_i$ is assigned to a single cluster $S_k$. Taking into account that the dispersion of feature $v$ in the cluster $S_k$ is given by $D_{kv}=\sum_{i \in S_k}|y_{iv} - c_{kv}|^p$, we can apply the first-order optimality condition for the Lagrange function $L$ as follows:
\begin{equation*}
L=\sum_{v \in V} w_{kv}^p D_{kv} + \lambda\left(1 - \sum_{v \in V} w_{kv}\right).
\end{equation*}
The derivative of $L$ with respect to the feature weight $w_{kv}$ is given by
\begin{equation*}
\frac{\partial L}{\partial w_{kv}}=p w_{kv}^{p-1} D_{kv} - \lambda.
\end{equation*}
By equating this to zero, we find 
\begin{equation*}
\left(\frac{\lambda}{p}\right)^{\frac{1}{p-1}}=w_{kv} D_{kv}^{\frac{1}{p-1}},
\end{equation*}
\begin{equation*}
w_{kv}=\left(\frac{\lambda}{pD_{kv}}\right)^{\frac{1}{p-1}}.
\end{equation*}
Summing the above over all $v \in V$,
\begin{equation*}
L=\sum_{v \in V}\left(\frac{\lambda}{pD_{kv}}\right)^{\frac{1}{p-1}},
\end{equation*}
so that 
\begin{equation*}
\left(\frac{\lambda}{p}\right)^{\frac{1}{p-1}}=\frac{1}{\sum_{v \in V}(\frac{1}{D_{kv}})^{\frac{1}{p-1}}}.
\end{equation*}
The above leads to (\ref{Eq:FeatureWeights}) unless the dispersion of a given feature $v$ in a cluster $S_k$ is equal to zero, $D_{kv}=0$, which is avoided as described below (\ref{Eq:FeatureWeights}).

\subsection{iMWK-Means}\label{SreiMWK}

We use the principle of the iK-Means algorithm with distance (\ref{Eq:AdjustedDistanceMeasure}) for providing MWK-Means with good initial centroids and initial weights. This proved to be a good choice in comparative experiments \citep{de2012initializations}. We refer to this as the intelligent Minkowski Weighted K-Means (iMWK-Means) \citep{cordeiro2011minkowski}. Again, this requires a tuning constant $\theta$, interpreted as before. In our experiments we have set $\theta$ to one (see Section \ref{Sexp}).
\begin{enumerate}
\label{Alg:iMWK}
\itemsep0em 
\item Standardize the data in $Y$.
\item Set $c_c$, to be the Minkowski centre of Y, and each $w_{kv}=1/|V|$.
\item Set $c_t=\argmax_{y_i} d_p(y_i,c_c)$, the farthest observation $y_i \in Y$ from $c_c$ according to (\ref{Eq:AdjustedDistanceMeasure}).
\item Run MWK-Means using $c_t$ and $c_c$ as initial centroids, forming clusters $S_c$ and $S_t$ without moving $c_c$, i.e. in step 3 of MWK-Means, $c_c$ is kept fixed.
\item If $|S_t|\ge \theta$ add $c_t$ to $C_{init}$ and the weight $w$ yielded by MWK-Means to $W_{init}$. In any case, remove the entities $y_i \in S_t$ from $Y$.
\item If there are still any entities to be clustered, go to Step 2.
\item Run MWK-Means on the original data set $Y$ initialized with the centroids in $C_{init}$, the initial Weights in $W_{init}$ and set $K=|C_{init}|$.
\end{enumerate}
If the desired value of $K$ is known, the algorithm above can still be used to find good initial centroids for MWK-Means. In this case one can set $\theta$ to one, forcing this method to recover all anomalous clusters in $Y$. $K$ initial centroids are then found by removing from $C_{init}$ and $W_{init}$ all but the $K$ entries related to the anomalous clusters with the highest number of entities.

Instead of taking $K=|C_{init}|$ at face value, we set $\theta$ to one and run iMWK-Means once for each $K \in\{2,\ldots,min(|C_{init}|,20)\}$. We then estimate the optimal $K$ using each of the cluster validity indexes presented in Section \ref{Sec:Background}.

In our experiments, we have decided not to use the output of iMWK-Means given by (\ref{Eq:MWK-Means}) for the Hartigan and Calinski and Harabasz indexes. These two clustering validity indexes were designed specifically for the squared Euclidean metric and are not easily generalized to other Minkowski metrics, so we supply them instead with the square Euclidean sum of distances between entities and their respective centroids estimated by iMWK-Means.

Our iMWK-Means algorithm is clearly based on K-Means. The computation time of latter at a given iteration is proportional to the product of the number of entities and features. While these two numbers are known the number of iterations is not, and it can be rather high in the worst-case scenario \citep{arthur2006worst}. However, the number of iterations necessary for convergence can be considerably reduced should K-Means have good initial centroids, such as those generated by the iK-Means \citep{de2013empirical}. Previously, we performed a series of experiments recording the CPU running time for various clustering algorithms including K-Means, MWK-Means and iMWK-Means \citep{cordeiro2011minkowski}. As one would expect, K-Means normally converges faster than iMWK-Means. However, K-Means is a non-deterministic algorithm normally ran a number of times per data set. Running K-Means 100 times on data sets containing spherical Gaussian clusters and a high proportion of noise features would, in average, take longer than running iMWK-Means.

\subsection{iMWK-Means with explicit re-scaling}\label{SreiMWKr}

Our second method is a variation of the first one, this time with an explicit re-scaling of the data set and the centroids, specifically for estimating $K$, using the following algorithm:
\begin{enumerate}
\itemsep0em 
\item Run iMWK-Means. For $K \in \{2,\ldots,min(|C_{init}|,20)\}$:
\begin{enumerate}
\item 	 Generate $Y_w$ and $C_w$ as specified below.
\item    Apply a given CVI using $Y_w$, $C_w$ and the clustering generated by iMWK-Means with $K$ fixed as explained in Section \ref{SreiMWK}.
\item    Increment $K$ by one.
\end{enumerate}
\end{enumerate}
$Y_w$ denotes the re-scaled version of the data set $Y$. We represent the value of a feature $v$ in a particular entity $y_i$ in $Y_w$ by $y_{ivw}$. Given clustering $S=\{S_1, S_2, ..., S_K\}$ produced by iMWK-Means at a given $K$, we obtain $Y_w$ by setting each $y_{ivw} \in S_k$ to $y_{iv} w_{kv}$, for each cluster $S_k \in S$ (to be understood as including the initial standardisation).

Similarly, $C_w$ contains the re-scaled centroids $c_w$ defined by $c_{kvw}=c_{kv}w_{kv}$ for $c_k\in C$. Our data re-scaling process takes into account the relative dispersion of each feature $v$ in each cluster $S_k$. We find this to be a considerably better way to re-scale features than using equations that treat all features equally regardless of the data structure, like the popular \textit{z}-score.

As for our previous method, the Hartigan and CH indexes are supplied with the square Euclidean sum of distances between entities and their respective centroids, rather than the output of the iMWK-Means criterion given by Equation (\ref{Eq:MWK-Means}).

\subsection{iMWK-Means with explicit re-scaling followed by K-Means}\label{SreiMWKrK}

Our third and last method takes the previous ideas even further. We now take into account that although iMWK-Means applies feature weights, these are not optimized in the beginning of the clustering process. This means that in most cases the optimal weights found by iMWK-Means are only used in the clustering in its very last iteration, while suboptimal weights are used in all previous iterations. For this reason we propose to re-cluster $Y_w$ with the final weights from iMWK-Means using a fully iterated K-Means (which has the additional advantage that the K-Means based indexes Hartigan and CH apply in a more natural manner than before).

\begin{enumerate}
\itemsep0em 
\item Run iMWK-Means. For $K \in \{2,\ldots,min(|C_{init}|,20)\}$:
\begin{enumerate}
\item 	 Generate $Y_w$.
\item 	 Apply K-Means to $Y_w$ from random initialisations
$q$ times ($q=100$ here), disregarding all clusterings but the one returning thesmallest K-Means criterion output (\ref{Eq:K_Means}).
\item    Apply a given CVI using $Y_w$, as well as the clustering generated by K-Means.
\item    Increment $K$ by one.
\end{enumerate}
\end{enumerate}

We note that although we are using K-Means in the re-clustering stage of the above method, we could indeed have chosen any other clustering algorithm. 

\subsection{Summary of methods}
Summarizing, MWK- and iMWK-Means (Section \ref{SreiMWK})
cluster the data according to (\ref{Eq:MWK-Means}), but the raw (standardised) data are used for computing the CVIs and thus for estimating $K$. 

The following two new methods are aimed at making a better use of the variable re-scaling that implicitly takes place in iMWK-Means.  
iMWK-Means with explicit re-scaling (Section \ref{SreiMWKr}) uses the iMWK-Means clustering, but evaluates the CVIs with the re-scaled data that was implicitly used by (\ref{Eq:MWK-Means}) for clustering. The idea is that if the re-scaling delivers a better clustering, it can also be expected to deliver a better estimate of $K$.

iMWK-Means with explicit re-scaling followed by K-Means (Section \ref{SreiMWKrK}) uses iMWK-Means as technique for re-scaling rather than as clustering technique, and computes K-Means on the re-scaled data, which is then fed into the CVIs. The idea is that the major benefit of MWK-Means compared with K-Means may actually be the re-scaling of the variables, but (\ref{Eq:K_Means}) applied to the re-scaled data may be the more appropriate clustering criterion; at least this is certainly the case when used together with the CVIs that are specifically based on (\ref{Eq:K_Means}), namely Hartigan and CH.

\section{Simulations}\label{Sexp}
\subsection{Standard scenarios}\label{ssimstan}
For the experiments in this paper we have generated a total of 600 synthetic data sets. We have used 12 different configurations, with and without noise features. The noise features are comprised of uniformly random values within the domain of the data set. These are pseudo-random values drawn from the standard uniform distribution. Each data set has spherical Gaussian clusters with diagonal covariance matrices of $\sigma^2=0.5$. All centroid components were independently generated from the Gaussian distribution $N(0,1)$, and each point had a chance of $1/K$ to come from any cluster. 

We initially generated 50 data sets for each of the following configurations: (i) 1000 entities and 8 features partitioned into 2 clusters (1000x8-2); (ii) 1000 entities and 12 features partitioned into three clusters (1000x12-3); (iii) 1000 entities and 16 features partitioned into four clusters (1000x16-4); (iv) 1000 entities and 20 features partitioned into five clusters (1000x20-5), which defines a subtotal of 200 data sets.

We have then generated further 200 data sets by adding 50\% of features composed of uniformly random values to each configuration, and then another 200 data sets by adding 100\% of uniformly random features to the original data sets. Table \ref{Tab:ListOfDatasets} shows a list of all data sets we have used.   

\begin{table}[h]\small
\caption{Data sets, we generated 50 data sets under each of the following configurations}
\begin{center}
\begin{tabular}{lccccc}
&&Relevant&&Noise&Total Number\\
&Entities&Features&Clusters&features&of features\\
\cline{2-6}
1000x8-2&1000&8&2&0&8\\
1000x8-2 + 4NF&1000&8&2&4&12\\
1000x8-2 + 8NF&1000&8&2&8&16\\
\cline{1-6}
1000x12-3&1000&12&3&0&12\\
1000x12-3 + 6NF&1000&12&3&6&18\\
1000x12-3 + 12NF&1000&12&3&12&24\\
\cline{1-6}
1000x16-4&1000&16&4&0&16\\
1000x16-4 + 8NF&1000&16&4&8&24\\
1000x16-4 + 16NF&1000&16&4&16&32\\
\cline{1-6}
1000x20-5&1000&20&5&0&20\\
1000x20-5 + 10NF&1000&20&5&10&30\\
1000x20-5 + 20NF&1000&20&5&20&40\\
\end{tabular}
\end{center}
\label{Tab:ListOfDatasets}
\end{table}


%

We have run three sets of experiments, each based on the number of noise features added to a data set, none, 50\%, and 100\% extra noise features, as explained above. In each of these we run experiments with our three methods (details in Section \ref{Sec:RescalingMethod}), as well as experiments with K-Means without re-scaling the data in order to create a baseline, which we explain in the following. We call these experiments ``standard scenarios''; see Section \ref{ssimother} for experiments with t-distributions and correlation.

Aiming to achieve a feasible time consumption and a realistic simulation we have decided to search for $K$ in the interval $[2,20]$, although this can still be computationally demanding for large data sets. Regardless, we ran K-Means 100 times for any given standardized data set $Y$, and applied each of the CVIs in Section \ref{Sec:Background} to the clustering with the lowest output criterion given by (\ref{Eq:K_Means}), for each $K=2, 3, ..., 20$. We then select $K$ by optimizing each CVI.

In order to adapt the CVIs appropriately to the K-Means method, we used some modifications in some of the baseline experiments, in which K-Means was ran using the squared Euclidean distance, with the exception of the experiments applying the Silhouette using the Manhattan distance. With the latter we ran K-Means with the Manhattan distance, leading to the well established K-Medians \citep{jain1988algorithms} as baseline. Second, for experiments that are referred to as using the Minkowski distance for the Silhouette and Dunn's indexes, in the baseline experiments we used $p=2$, so that these indexes are aligned with the standard K-Means clustering algorithm, making them equivalent to the Euclidean versions, whereas for the experiments with re-scaling methods we used the Minkowski-distance with flexible $p$, as now explained. 

Our experiments with our re-scaling methods presented in Section \ref{Sec:RescalingMethod} allow a further reduction in the interval used to search for $K$ thanks to iMWK-Means use of a $L_p$ version of iK-Means anomalous pattens (as explained in Section \ref{Sec:FeatureWeighting}). By setting  $\theta$ to $1$ we are able to determine the maximum number of clusters a given data set should have ($|C_{init}|$). Such a low value for $\theta$ allows for the acceptance of clusters of any cardinality, including singletons. When searching for $K$ using one of our feature re-scaling methods, we have done so in the interval $[2, min(|C_{init}|, 20)]$ as explained above.

Our experiments have three general objectives. First, They allow us to analyse the changes in the relative error in the estimation of $K$ given by each CVI we experiment with:
\begin{equation}
RE=\frac{|K-K_{Est}|}{K},
\end{equation}
where $K_{Est}$ is the estimated number of clusters. 

Second, the experiments allow us to quantify the number of times each clustering validity index estimates the exact number of clusters in the data set, which is the ultimate objective of these indexes.

Third, we can see if the estimated values of $K$ actually return good clusterings. We can measure this by using the adjusted Rand index (ARI) given by their respective clusterings:
\begin{equation}
ARI = \frac{ \sum_{ij} \binom{n_{ij}}{2} - [\sum_i \binom{a_i}{2} \sum_j \binom{b_j}{2}] / \binom{n}{2} }{ \frac{1}{2} [\sum_i \binom{a_i}{2} + \sum_j \binom{b_j}{2}] - [\sum_i \binom{a_i}{2} \sum_j \binom{b_j}{2}] / \binom{n}{2} },
\end{equation}
where $n_{ij}=|S_i \cap S_j|$, $a_i = \sum_{j=1}^K |S_i \cap S_j|$ and $b_i = \sum_{i=1}^K |S_i \cap S_j|$.
\subsection{Results and discussion}\label{Sec:Results}

Our first set of experiments addresses the data sets with no noise features added to them. Since these are data sets with equally relevant features we find it reasonable to expect good results from the cluster validity indexes even if we do not apply any of the methods we introduced in Section \ref{Sec:RescalingMethod}. 
Tables \ref{Tab:RelativeErrorNoNoise}, \ref{Tab:ARINoNoise} and \ref{Tab:PercentFindRightClusters_NoNoise}, show the relative error, the adjusted rand index, and the percentage of times each index found the true number of clusters, respectively. Each table presents the results we obtained by clustering the data sets using K-Means (our baseline), iMWK-Means, iMWK-Means with explicit re-scaling and iMWK-Means with explicit re-scaling followed by K-Means in relation to the known number of clusters in each data set. 

The values of $p$ in each table relate to the $p$ used in iMWK-Means. For this reason we do not present experiments for different $p$ values for our baseline experiment as these do not use iMWK-Means, but only K-Means. Our tables also present two cluster validity indexes that can incorporate the Minkowski distance (these have been marked with \textit{Mink}). These two indexes use the same value of $p$ as iMWK-Means. For instance, in Table \ref{Tab:RelativeErrorNoNoise} the entries under $p=1.4$ specify that iMWK-Means was ran with this particular $p$. The results presented for the Silhouette (Mink) and Dunn (Mink) in row $p=1.4$ also use the same $p=1.4$ for the distances used within the index calculation.

In each table the row labelled $p\rightarrow 1$, presents the results for a $p$ very close to $1$, $1.00001$. We have not experimented with $p=1$ because in this scenario the minimization of (\ref{Eq:FeatureWeights}) sets a single feature weight to one and all others to zero \citep{cordeiro2011minkowski}. We believe that using a single feature is very unlikely to produce good results and do not follow this path here.

Table \ref{Tab:RelativeErrorNoNoise} shows that in most cases explicit re-scaling does not seem to reduce the relative error in relation to the baseline provided by K-Means. However, one could claim this is an expected behaviour since in these data sets each feature $v \in V$ has about the same degree of relevance. While the K-Means baseline had the best results for Silhouette (Euclidean and Manhattan), our iMWK-Means based method had the best results for Silhouette (Minkowski), Dunn's (Minkowski), CH as well as the Hartigan index. We find it interesting to see that the  lowest relative error overall was given by the Silhouette using the Minkowski distance at $p=2$; this may be related to the fact that these are well separated Euclidean clusters.

Table \ref{Tab:ARINoNoise}, which presents the adjusted Rand index, shows in a considerably clearer way that $p=2$ is indeed the best distance exponent. Our iMWK-Means based method at $p=2$ provides the highest adjusted Rand index for five of the seven cluster validity indexes we experiment with, including the best overall adjusted rand index given by the Silhouette index using either Euclidean or Minkowski distances (at $p=2$ these are equivalent).

Table \ref{Tab:PercentFindRightClusters_NoNoise} shows the percentage of times each index estimated the true number of clusters. We can clearly see that iMWK-Means at $p=2$ generates the best and second best overall results of 91\% (Silhouette, Euclidean) and 88.5\% (Silhouette, Manhattan). Although $p=2$ did not provide the best results for Dunn's under the Euclidean and Minkowski distances, we can still see that at $p=2$ there is a considerable improvement of 9\%. At $p=2$ CH presents an improvement of 0.5\% (the best at $p=1.4$ is 1\%). 


%
\begin{table}[h]\scriptsize
\caption{Relative errors of the estimation of the number of clusters and their standard errors in GM data with no noise in relation to the known number of classes.}
\begin{center}
\tabcolsep=0.04cm 
\begin{tabular}{@{\extracolsep{2pt}}lccccccc@{}}
&\multicolumn{3}{c}{Silhouette}&\multicolumn{2}{c}{Dunn}\\
\cline{2-4}
\cline{5-6}
&Eucl&Manh&Mink&Eucl&Mink&CH&Hartigan\\
\cline{2-8}
K-Means&\textbf{0.030}/0.01&\textbf{0.036}/0.01&0.030/0.01&0.483/0.12&0.483/0.12&0.045/0.01&3.300/0.15\\ 
\cline{1-8}
\multicolumn{8}{l}{iMWK-Means}\\
$p\rightarrow 1$&0.301/0.02&0.300/0.02&0.300/0.02&0.358/0.03&0.395/0.03&0.282/0.02&0.618/0.05\\ 
$p=1.1$&0.287/0.02&0.286/0.02&0.286/0.02&0.303/0.02&0.305/0.02&0.289/0.02&0.664/0.05\\ 
$p=1.2$&0.096/0.01&0.100/0.01&0.100/0.01&0.119/0.01&0.114/0.01&0.104/0.01&0.686/0.05\\
$p=1.3$&0.046/0.01&0.050/0.01&0.048/0.01&0.068/0.01&0.055/0.01&0.050/0.01&0.517/0.05\\ 
$p=1.4$&0.035/0.01&0.041/0.01&0.038/0.01&0.059/0.01&\textbf{0.051}/0.01&\textbf{0.042}/0.01&\textbf{0.485}/0.05\\ 
$p=1.5$&0.036/0.01&0.043/0.01&0.039/0.01&0.057/0.01&\textbf{0.051}/0.01&0.046/0.01&0.608/0.06\\ 
$p=1.6$&0.036/0.01&0.042/0.01&0.037/0.01&0.059/0.01&0.054/0.01&0.044/0.01&0.759/0.06\\ 
$p=1.7$&0.037/0.01&0.043/0.01&0.038/0.01&0.067/0.01&0.064/0.01&0.046/0.01&0.909/0.08\\ 
$p=1.8$&0.034/0.01&0.040/0.01&0.034/0.01&0.066/0.02&0.062/0.02&0.046/0.01&0.892/0.08\\ 
$p=1.9$&0.033/0.01&0.042/0.01&0.033/0.01&0.070/0.01&0.070/0.01&0.047/0.01&0.962/0.08\\ 
$p=2$&0.029/0.01&0.039/0.01&\textbf{0.029}/0.01&0.054/0.01&0.054/0.01&0.044/0.01&1.043/0.09\\ 
$p=2.5$&0.033/0.01&0.043/0.01&0.033/0.01&0.097/0.03&0.067/0.01&0.046/0.01&1.081/0.10\\ 
$p=3$&0.032/0.01&0.042/0.01&0.033/0.01&0.167/0.06&0.164/0.05&0.045/0.01&1.311/0.11\\ 
\cline{1-8}
\multicolumn{8}{l}{iMWK-Means with explicit re-scaling}\\
$p\rightarrow 1$&1.152/0.09&1.132/0.09&1.132/0.09&1.095/0.09&1.112/0.09&0.902/0.07&0.974/0.07\\ 
$p=1.1$&0.698/0.06&1.329/0.10&1.301/0.10&0.320/0.02&0.318/0.02&1.468/0.08&1.128/0.08\\ 
$p=1.2$&0.559/0.09&0.941/0.11&0.928/0.11&0.129/0.01&0.136/0.01&1.579/0.09&1.178/0.08\\ 
$p=1.3$&0.221/0.06&0.327/0.08&0.355/0.09&0.067/0.01&0.086/0.01&1.161/0.11&1.411/0.10\\ 
$p=1.4$&0.062/0.01&0.061/0.01&0.106/0.05&0.059/0.01&0.071/0.01&0.577/0.10&1.356/0.10\\ 
$p=1.5$&0.059/0.01&0.057/0.01&0.059/0.01&0.072/0.02&0.087/0.02&0.134/0.03&1.393/0.10\\ 
$p=1.6$&0.057/0.01&0.054/0.01&0.055/0.01&0.152/0.05&0.097/0.02&0.108/0.01&1.355/0.09\\ 
$p=1.7$&0.051/0.01&0.052/0.01&0.053/0.01&0.145/0.04&0.146/0.04&0.106/0.01&1.336/0.10\\ 
$p=1.8$&0.051/0.01&0.051/0.01&0.051/0.01&0.122/0.03&0.125/0.03&0.104/0.01&1.560/0.12\\ 
$p=1.9$&0.050/0.01&0.049/0.01&0.050/0.01&0.158/0.04&0.143/0.04&0.100/0.01&1.451/0.11\\ 
$p=2$&0.046/0.01&0.045/0.01&0.046/0.01&0.200/0.06&0.200/0.06&0.096/0.01&1.518/0.11\\ 
$p=2.5$&0.046/0.01&0.047/0.01&0.046/0.01&0.182/0.06&0.201/0.06&0.075/0.01&1.522/0.13\\ 
$p=3$&0.043/0.01&0.047/0.01&0.046/0.01&0.212/0.06&0.240/0.07&0.072/0.01&1.845/0.15\\ 
\cline{1-8}
\multicolumn{8}{l}{iMWK-Means with explicit re-scaling followed by K-Means}\\
$p\rightarrow 1$&1.280/0.10&1.095/0.09&1.173/0.10&1.365/0.09&1.350/0.09&1.183/0.08&0.936/0.06\\ 
$p=1.1$&0.734/0.07&0.989/0.08&0.943/0.08&0.314/0.02&0.318/0.02&2.063/0.10&0.990/0.07\\ 
$p=1.2$&0.789/0.10&0.295/0.06&0.428/0.08&0.126/0.01&0.126/0.01&2.236/0.10&1.109/0.09\\ 
$p=1.3$&0.177/0.05&0.088/0.01&0.120/0.04&0.053/0.01&0.066/0.01&1.834/0.13&1.142/0.08\\ 
$p=1.4$&0.062/0.01&0.061/0.01&0.061/0.01&\textbf{0.048}/0.01&0.064/0.01&1.027/0.14&1.268/0.09\\ 
$p=1.5$&0.059/0.01&0.057/0.01&0.060/0.01&0.070/0.01&0.090/0.02&0.270/0.07&1.357/0.10\\ 
$p=1.6$&0.057/0.01&0.052/0.01&0.055/0.01&0.116/0.03&0.109/0.03&0.108/0.01&1.634/0.11\\ 
$p=1.7$&0.051/0.01&0.050/0.01&0.053/0.01&0.108/0.03&0.073/0.02&0.106/0.01&1.658/0.11\\ 
$p=1.8$&0.051/0.01&0.048/0.01&0.051/0.01&0.153/0.05&0.174/0.05&0.105/0.01&1.794/0.12\\ 
$p=1.9$&0.047/0.01&0.044/0.01&0.046/0.01&0.175/0.05&0.186/0.06&0.101/0.01&1.825/0.10\\ 
$p=2$&0.045/0.01&0.044/0.01&0.045/0.01&0.225/0.07&0.225/0.07&0.096/0.01&1.770/0.11\\ 
$p=2.5$&0.045/0.01&0.046/0.01&0.045/0.01&0.375/0.10&0.358/0.10&0.075/0.01&2.067/0.13\\ 
$p=3$&0.041/0.01&0.045/0.01&0.044/0.01&0.505/0.12&0.346/0.09&0.071/0.01&2.379/0.15\\ 
\end{tabular}
\end{center}
\label{Tab:RelativeErrorNoNoise}
\end{table}
%
%
%

\begin{table}[h]\footnotesize
\caption{The adjusted Rand index and its standard error for the clusterings at each estimated $K$ in the data sets with no noise features.}
\begin{center}
\tabcolsep=0.04cm 
\begin{tabular}{@{\extracolsep{2pt}}lccccccc@{}}
&\multicolumn{3}{c}{Silhouette}&\multicolumn{2}{c}{Dunn}\\
\cline{2-4}
\cline{5-6}
&Eucl&Manh&Mink&Eucl&Mink&CH&Hartigan\\
\cline{2-8}
K-Means&0.951/0.01&0.940/0.01&0.951/0.01&0.861/0.02&0.861/0.02&0.937/0.01&0.363/0.01\\
\cline{1-8}
\multicolumn{6}{l}{iMWK-Means}\\
$p\rightarrow 1$&0.459/0.02&0.459/0.02&0.459/0.02&0.431/0.02&0.429/0.02&0.468/0.02&0.337/0.01\\
$p=1.1$&0.619/0.02&0.617/0.02&0.617/0.02&0.601/0.02&0.598/0.02&0.615/0.02&0.374/0.01\\
$p=1.2$&0.873/0.01&0.868/0.01&0.868/0.01&0.846/0.02&0.853/0.01&0.869/0.01&0.542/0.02\\
$p=1.3$&0.933/0.01&0.927/0.01&0.929/0.01&0.905/0.01&0.922/0.01&0.930/0.01&0.760/0.02\\
$p=1.4$&0.947/0.01&0.940/0.01&0.944/0.01&0.916/0.01&0.926/0.01&0.941/0.01&\textbf{0.799}/0.02\\
$p=1.5$&0.946/0.01&0.939/0.01&0.943/0.01&0.919/0.01&0.926/0.01&0.938/0.01&0.786/0.02\\
$p=1.6$&0.947/0.01&0.940/0.01&0.946/0.01&0.921/0.01&\textbf{0.927}/0.01&0.940/0.01&0.755/0.02\\
$p=1.7$&0.946/0.01&0.939/0.01&0.945/0.01&0.921/0.01&0.925/0.01&0.938/0.01&0.731/0.02\\
$p=1.8$&0.950/0.01&\textbf{0.943}/0.01&0.950/0.01&0.923/0.01&\textbf{0.927}/0.01&0.938/0.01&0.746/0.02\\
$p=1.9$&0.950/0.01&0.940/0.01&0.950/0.01&0.918/0.01&0.919/0.01&0.936/0.01&0.723/0.02\\
$p=2$&\textbf{0.955}/0.01&\textbf{0.943}/0.01&\textbf{0.955}/0.01&0.927/0.01&\textbf{0.927}/0.01&\textbf{0.939}/0.01&0.715/0.02\\
$p=2.5$&0.950/0.01&0.939/0.01&0.950/0.01&0.916/0.01&0.918/0.01&0.937/0.01&0.709/0.02\\
$p=3$&0.950/0.01&0.938/0.01&0.948/0.01&0.898/0.01&0.878/0.01&0.937/0.01&0.687/0.02\\
\cline{1-8}
\multicolumn{6}{l}{iMWK-Means with explicit re-scaling}\\
$p\rightarrow 1$&0.330/0.02&0.329/0.02&0.329/0.02&0.306/0.02&0.306/0.02&0.346/0.02&0.285/0.01\\
$p=1.1$&0.504/0.02&0.401/0.02&0.410/0.02&0.595/0.02&0.594/0.02&0.290/0.01&0.364/0.01\\
$p=1.2$&0.724/0.02&0.647/0.02&0.645/0.02&0.840/0.02&0.830/0.02&0.410/0.02&0.506/0.02\\
$p=1.3$&0.862/0.02&0.849/0.02&0.851/0.02&0.911/0.01&0.899/0.01&0.651/0.02&0.581/0.02\\
$p=1.4$&0.915/0.01&0.915/0.01&0.912/0.01&0.925/0.01&0.909/0.01&0.790/0.02&0.661/0.02\\
$p=1.5$&0.921/0.01&0.921/0.01&0.919/0.01&0.920/0.01&0.912/0.01&0.866/0.01&0.669/0.02\\
$p=1.6$&0.922/0.01&0.925/0.01&0.924/0.01&0.911/0.01&0.913/0.01&0.871/0.01&0.664/0.02\\
$p=1.7$&0.929/0.01&0.927/0.01&0.927/0.01&0.903/0.01&0.904/0.01&0.872/0.01&0.671/0.02\\
$p=1.8$&0.930/0.01&0.929/0.01&0.929/0.01&0.907/0.01&0.903/0.01&0.875/0.01&0.658/0.02\\
$p=1.9$&0.930/0.01&0.931/0.01&0.930/0.01&0.900/0.01&0.902/0.01&0.879/0.01&0.665/0.02\\
$p=2$&0.935/0.01&0.936/0.01&0.935/0.01&0.896/0.01&0.896/0.01&0.883/0.01&0.647/0.02\\
$p=2.5$&0.935/0.01&0.934/0.01&0.936/0.01&0.898/0.01&0.899/0.01&0.905/0.01&0.665/0.02\\
$p=3$&0.938/0.01&0.933/0.01&0.934/0.01&0.887/0.01&0.879/0.02&0.907/0.01&0.642/0.02\\
\cline{1-8}
\multicolumn{8}{l}{iMWK-Means with explicit re-scaling followed by K-Means}\\
$p\rightarrow 1$&0.294/0.02&0.314/0.02&0.303/0.02&0.236/0.01&0.244/0.02&0.275/0.01&0.248/0.01\\ 
$p=1.1$&0.400/0.02&0.413/0.02&0.408/0.02&0.543/0.02&0.542/0.02&0.210/0.01&0.331/0.02\\ 
$p=1.2$&0.670/0.02&0.768/0.02&0.762/0.02&0.842/0.02&0.840/0.02&0.293/0.01&0.494/0.02\\ 
$p=1.3$&0.873/0.01&0.880/0.01&0.886/0.01&0.932/0.01&0.915/0.01&0.510/0.02&0.609/0.02\\ 
$p=1.4$&0.915/0.01&0.913/0.01&0.916/0.01&\textbf{0.940}/0.01&0.925/0.01&0.715/0.02&0.627/0.02\\ 
$p=1.5$&0.921/0.01&0.918/0.01&0.919/0.01&0.924/0.01&0.915/0.01&0.850/0.02&0.614/0.02\\ 
$p=1.6$&0.923/0.01&0.924/0.01&0.924/0.01&0.911/0.01&0.913/0.01&0.871/0.01&0.570/0.02\\ 
$p=1.7$&0.929/0.01&0.926/0.01&0.927/0.01&0.909/0.01&0.918/0.01&0.873/0.01&0.578/0.02\\ 
$p=1.8$&0.930/0.01&0.928/0.01&0.929/0.01&0.905/0.01&0.905/0.01&0.874/0.01&0.563/0.02\\ 
$p=1.9$&0.934/0.01&0.933/0.01&0.934/0.01&0.902/0.01&0.901/0.01&0.878/0.01&0.541/0.02\\ 
$p=2$&0.936/0.01&0.932/0.01&0.936/0.01&0.895/0.01&0.895/0.01&0.883/0.01&0.563/0.02\\ 
$p=2.5$&0.936/0.01&0.930/0.01&0.936/0.01&0.878/0.02&0.884/0.02&0.905/0.01&0.539/0.02\\ 
$p=3$&0.941/0.01&0.931/0.01&0.937/0.01&0.865/0.02&0.876/0.02&0.909/0.01&0.507/0.02\\ 
\end{tabular}
\end{center}
\label{Tab:ARINoNoise}
\end{table}
%
%
%

\begin{table}[h]\footnotesize
\caption{The percentage of finding the true number of clusters in data sets with no noise features.}
\begin{center}
\tabcolsep=0.04cm 
\begin{tabular}{@{\extracolsep{2pt}}lccccccc@{}}
&\multicolumn{3}{c}{Silhouette}&\multicolumn{2}{c}{Dunn}\\
\cline{2-4}
\cline{5-6}
&Eucl&Manh&Mink&Eucl&Mink&CH&Hartigan\\
\cline{2-8}
K-Means&\textbf{89.500}&88.000&89.500&75.500&75.500&87.000&1.000\\ 
\cline{1-8}
\multicolumn{6}{l}{iMWK-Means}\\
$p\rightarrow 1$&32.500&33.000&33.000&28.500&31.000&35.000&23.500\\ 
$p=1.1$&34.000&34.000&34.000&34.500&35.000&34.000&17.000\\ 
$p=1.2$&69.000&68.000&68.000&66.000&67.500&70.000&20.500\\ 
$p=1.3$&84.500&83.500&84.000&80.500&83.500&85.000&47.000\\ 
$p=1.4$&88.000&86.500&87.500&84.000&\textbf{85.500}&\textbf{88.000}&\textbf{49.500}\\ 
$p=1.5$&88.000&86.500&87.500&83.500&84.500&86.500&44.500\\ 
$p=1.6$&88.000&86.500&88.000&84.000&\textbf{85.500}&87.000&35.000\\ 
$p=1.7$&87.500&86.000&87.500&82.500&83.500&86.000&29.500\\ 
$p=1.8$&89.000&87.500&89.000&84.000&85.000&86.500&31.000\\ 
$p=1.9$&89.500&87.500&89.500&82.500&82.500&87.000&23.500\\ 
$p=2$&\textbf{91.000}&\textbf{88.500}&\textbf{91.000}&84.500&84.500&87.500&25.000\\ 
$p=2.5$&89.000&86.500&89.000&82.000&82.000&86.500&24.500\\ 
$p=3$&89.500&87.000&89.000&80.000&75.500&86.500&26.500\\ 
\cline{1-8}
\multicolumn{8}{l}{iMWK-Means with explicit re-scaling}\\
$p\rightarrow 1$&22.000&19.000&19.000&14.000&13.500&21.500&14.500\\ 
$p=1.1$&34.000&19.000&20.000&31.500&32.000&7.000&10.500\\ 
$p=1.2$&51.500&38.000&41.500&68.000&62.500&9.000&14.000\\ 
$p=1.3$&69.000&66.000&67.500&81.000&77.500&37.000&7.000\\ 
$p=1.4$&80.000&79.500&79.000&84.000&79.500&58.000&14.000\\ 
$p=1.5$&80.500&81.000&80.500&81.500&78.500&69.000&11.500\\ 
$p=1.6$&81.500&82.000&81.500&80.000&79.500&70.500&9.500\\ 
$p=1.7$&82.500&82.500&82.000&74.500&75.000&70.000&12.500\\ 
$p=1.8$&83.000&83.500&83.000&76.500&75.500&72.000&13.000\\ 
$p=1.9$&84.000&84.500&84.000&75.500&75.500&73.500&14.500\\ 
$p=2$&85.500&86.000&85.500&73.000&73.000&73.500&13.500\\ 
$p=2.5$&85.000&85.000&85.000&74.500&74.000&79.000&17.500\\ 
$p=3$&86.000&85.000&85.000&74.000&73.000&80.000&16.000\\ 
\cline{1-8}
\multicolumn{8}{l}{iMWK-Means with explicit re-scaling followed by K-Means}\\
$p\rightarrow 1$&19.000&19.500&19.000&10.500&11.500&12.000&10.000\\ 
$p=1.1$&29.500&21.500&23.000&32.500&32.500&2.500&13.500\\ 
$p=1.2$&46.500&49.500&51.000&69.500&69.500&4.000&14.500\\ 
$p=1.3$&70.500&70.500&72.500&85.500&80.500&26.000&22.000\\ 
$p=1.4$&80.000&79.500&79.500&\textbf{88.000}&83.000&50.000&18.500\\ 
$p=1.5$&80.500&81.000&80.000&83.500&81.000&67.000&16.500\\ 
$p=1.6$&81.500&82.500&81.500&82.000&81.500&70.500&13.500\\ 
$p=1.7$&82.500&83.000&82.000&80.500&82.500&70.000&18.500\\ 
$p=1.8$&83.000&84.000&83.000&79.500&79.500&71.500&16.000\\ 
$p=1.9$&84.500&85.500&84.500&78.500&78.500&73.000&14.000\\ 
$p=2$&85.500&86.000&85.500&78.000&78.000&73.500&16.500\\ 
$p=2.5$&85.000&85.000&85.000&76.500&78.000&79.000&15.000\\ 
$p=3$&86.500&85.500&85.500&75.000&76.500&80.500&14.000\\ 
\end{tabular}
\end{center}
\label{Tab:PercentFindRightClusters_NoNoise}
\end{table}

In our second set of experiments, we have added 50\% of noise features to each of our data sets. In this scenario we can see a different pattern in which the best results are given by our methods that perform explicit re-scaling of features. Tables \ref{Tab:RelativeErrorLowNoise}, \ref{Tab:ARILowNoise}, \ref{Tab:PercentFindRightClusters_LowNoise} show the results for this set of experiments regarding the relative error, the adjusted Rand index, and the percentage of times the algorithm found the true number of clusters.

Table \ref{Tab:RelativeErrorLowNoise} shows a clear impact of the noise features in our K-Means baseline. The Silhouette index (Euclidean), which was very successful in previous experiments, presents now a relative error that is more than 10 times higher than in our experiments with no noise features.

The method producing the best results is now iMWK-Means with explicit re-scaling followed by K-Means; results are clearly worse without applying K-Means again in the end. iMWK-Means with explicit re-scaling followed by K-Means presents the smallest overall relative error of $0.052$ at $p=1.4$ using the Silhouette and the Minkowski distance. The very same $p$ produced the best results for the Silhouette index using the Euclidean distance, $0.053$, and the Dunn's index using the Minkowski distance, $0.054$. Although $p=1.4$ did not produce the best relative error for the other cluster validation indexes, it produced results much better than the K-Means baseline for all other indexes. Most results for $p=1.3$ are substantially worse than for $p=1.4$, whereas choosing $p$ a bit larger than 1.4 does not affect the performance that strongly.

Table \ref{Tab:ARILowNoise} shows that the highest adjusted Rand index of $0.950$ is reached by iMWK-Means and the Hartigan index. The iMWK-Means with explicit re-scaling followed by K-Means using the Dunn index with the Minkowski distance is, at $0.936$, quite close to the maximum found. In both cases the best results are found at $p=1.4$. iMWK-Means with explicit rescaling but without the further K-Means step yields some good results with $p=1.7$, particularly with the Silhouette index. For the better indexes, $p$ between 1.4 and 1.9  yields good results.

Table \ref{Tab:PercentFindRightClusters_LowNoise} presents the percentage of times each of the cluster validation indexes estimates the true number of clusters in the data sets with 50\% extra noise features. Here we can see that the best results for all indexes except Hartigan is given by iMWK-Means with explicit re-scaling followed by K-Means, most at $p=1.4$ including the best overall result of 85.5\% using Dunn's index and the Minkowski distance.

 Out of the baseline K-Means methods, for 50\% noise variables as opposed to the no-noise setups, the Silhouette index (Manhattan) is the best by some distance according to all three criteria. 

\begin{table}[h]\footnotesize
\caption{Relative errors of the estimation of the number of clusters and their standard error in GM data with extra 50\% noise features, in relation the the known number of classes.}
\begin{center}
\tabcolsep=0.04cm 
\begin{tabular}{@{\extracolsep{2pt}}lccccccc@{}}
&\multicolumn{3}{c}{Silhouette}&\multicolumn{2}{c}{Dunn}\\
\cline{2-4}
\cline{5-6}
&Eucl&Manh&Mink&Eucl&Mink&CH&Hartigan\\
\cline{2-8}
K-Means&0.316/0.08&0.067/0.01&0.316/0.08&1.820/0.19&1.820/0.19&0.310/0.02&3.482/0.15\\ 
\cline{1-8}
\multicolumn{8}{l}{iMWK-Means}\\
$p\rightarrow 1$&0.319/0.02&0.317/0.02&0.317/0.02&0.462/0.05&0.541/0.05&0.310/0.02&0.326/0.03\\ 
$p=1.1$&0.282/0.02&0.273/0.02&0.273/0.02&0.296/0.02&0.281/0.02&0.306/0.02&0.334/0.03\\ 
$p=1.2$&0.110/0.01&0.096/0.01&0.097/0.01&0.164/0.01&0.128/0.02&0.248/0.02&0.125/0.02\\ 
$p=1.3$&0.108/0.01&0.088/0.01&0.091/0.01&0.180/0.02&0.116/0.02&0.258/0.02&\textbf{0.096}/0.02\\ 
$p=1.4$&0.093/0.01&0.073/0.01&0.078/0.01&0.216/0.03&0.180/0.04&0.259/0.02&0.126/0.03\\ 
$p=1.5$&0.100/0.01&0.082/0.01&0.089/0.01&0.207/0.02&0.183/0.03&0.249/0.02&0.178/0.03\\ 
$p=1.6$&0.103/0.01&0.081/0.01&0.091/0.01&0.210/0.02&0.187/0.03&0.252/0.02&0.226/0.04\\ 
$p=1.7$&0.098/0.01&0.079/0.01&0.095/0.01&0.219/0.02&0.221/0.03&0.255/0.02&0.323/0.05\\ 
$p=1.8$&0.102/0.01&0.081/0.01&0.098/0.01&0.255/0.04&0.224/0.03&0.259/0.02&0.321/0.05\\ 
$p=1.9$&0.109/0.01&0.095/0.01&0.109/0.01&0.247/0.03&0.234/0.03&0.247/0.02&0.386/0.05\\ 
$p=2$&0.115/0.01&0.104/0.01&0.115/0.01&0.321/0.04&0.321/0.04&0.250/0.02&0.428/0.06\\ 
$p=2.5$&0.139/0.01&0.129/0.01&0.151/0.01&0.432/0.07&0.443/0.07&0.262/0.02&0.557/0.07\\ 
$p=3$&0.157/0.01&0.158/0.01&0.174/0.01&0.486/0.07&0.482/0.06&0.277/0.02&0.692/0.08\\ 
\cline{1-8}
\multicolumn{8}{l}{iMWK-Means with explicit re-scaling}\\
$p\rightarrow 1$&1.370/0.10&1.405/0.10&1.405/0.10&1.262/0.10&1.311/0.10&1.547/0.11&1.116/0.07\\ 
$p=1.1$&1.078/0.10&1.743/0.12&1.712/0.12&0.300/0.02&0.299/0.02&1.300/0.07&1.311/0.08\\ 
$p=1.2$&0.438/0.08&1.239/0.12&1.304/0.12&0.107/0.02&0.229/0.04&1.420/0.11&1.073/0.07\\ 
$p=1.3$&0.175/0.03&0.711/0.07&0.522/0.08&0.186/0.02&0.321/0.04&0.399/0.08&0.437/0.08\\ 
$p=1.4$&0.165/0.05&0.681/0.07&0.384/0.06&0.283/0.04&0.352/0.05&0.103/0.01&0.672/0.08\\ 
$p=1.5$&0.114/0.02&0.659/0.08&0.338/0.06&0.285/0.04&0.373/0.05&0.095/0.01&1.053/0.09\\ 
$p=1.6$&0.147/0.02&0.536/0.06&0.260/0.04&0.291/0.04&0.410/0.05&0.099/0.01&1.365/0.09\\ 
$p=1.7$&0.210/0.03&0.516/0.06&0.270/0.04&0.599/0.07&0.721/0.08&0.103/0.01&1.511/0.08\\ 
$p=1.8$&0.245/0.05&0.483/0.07&0.306/0.05&0.779/0.10&0.843/0.10&0.103/0.01&1.759/0.09\\ 
$p=1.9$&0.310/0.04&0.401/0.05&0.316/0.04&0.915/0.10&0.954/0.10&0.112/0.01&1.921/0.10\\ 
$p=2$&0.350/0.05&0.392/0.05&0.350/0.05&0.910/0.10&0.910/0.10&0.116/0.01&2.042/0.10\\ 
$p=2.5$&0.282/0.03&0.210/0.03&0.293/0.04&1.425/0.12&1.339/0.13&0.152/0.01&2.236/0.09\\ 
$p=3$&0.290/0.04&0.209/0.03&0.414/0.04&1.489/0.13&1.277/0.12&0.175/0.01&2.455/0.08\\ 
\cline{1-8}
\multicolumn{8}{l}{iMWK-Means with explicit re-scaling followed by K-Means}\\
$p\rightarrow 1$&1.609/0.11&1.318/0.10&1.512/0.11&1.566/0.10&1.527/0.10&1.766/0.10&1.174/0.07\\ 
$p=1.1$&1.217/0.10&1.188/0.10&1.263/0.11&0.304/0.02&0.298/0.02&2.438/0.10&1.355/0.09\\ 
$p=1.2$&0.488/0.09&0.213/0.06&0.219/0.06&0.077/0.01&0.082/0.01&2.578/0.11&1.270/0.09\\ 
$p=1.3$&0.073/0.01&0.074/0.01&0.070/0.01&0.086/0.02&0.076/0.01&1.289/0.14&1.046/0.09\\ 
$p=1.4$&\textbf{0.053}/0.01&0.057/0.01&\textbf{0.052}/0.01&0.079/0.02&\textbf{0.054}/0.01&0.282/0.08&1.373/0.09\\ 
$p=1.5$&0.056/0.01&0.063/0.01&0.057/0.01&0.067/0.01&0.056/0.01&\textbf{0.093}/0.01&1.634/0.09\\ 
$p=1.6$&0.060/0.01&0.061/0.01&0.062/0.01&\textbf{0.064}/0.01&0.067/0.01&0.095/0.01&1.698/0.08\\ 
$p=1.7$&0.058/0.01&0.062/0.01&0.060/0.01&0.138/0.04&0.130/0.03&0.100/0.01&2.013/0.10\\ 
$p=1.8$&0.066/0.01&0.067/0.01&0.074/0.01&0.110/0.03&0.138/0.03&0.101/0.01&2.203/0.10\\ 
$p=1.9$&0.074/0.01&0.077/0.01&0.078/0.01&0.134/0.03&0.136/0.03&0.110/0.01&2.153/0.10\\ 
$p=2$&0.090/0.02&0.113/0.03&0.090/0.02&0.214/0.06&0.214/0.06&0.109/0.01&2.248/0.09\\ 
$p=2.5$&0.097/0.01&0.059/0.01&0.097/0.01&0.483/0.10&0.474/0.10&0.139/0.01&2.524/0.11\\ 
$p=3$&0.097/0.01&\textbf{0.053}/0.01&0.112/0.01&0.623/0.12&0.535/0.10&0.158/0.01&2.848/0.11\\ 
\end{tabular}
\end{center}
\label{Tab:RelativeErrorLowNoise}
\end{table}
%
%
%

\begin{table}[h]\footnotesize
\caption{The adjusted Rand index and its standard error for the clusterings at each estimated $K$ in the data sets with about 50\% extra features composed of uniformly random noise.}
\begin{center}
\tabcolsep=0.04cm 
\begin{tabular}{@{\extracolsep{2pt}}lccccccc@{}}
&\multicolumn{3}{c}{Silhouette}&\multicolumn{2}{c}{Dunn}\\
\cline{2-4}
\cline{5-6}
&Eucl&Manh&Mink&Eucl&Mink&CH&Hartigan\\
\cline{2-8}
K-Means&0.817/0.02&0.897/0.01&0.817/0.02&0.578/0.02&0.578/0.02&0.580/0.02&0.335/0.01\\ 
\cline{1-8}
\multicolumn{6}{l}{iMWK-Means}\\
$p\rightarrow 1$&0.446/0.02&0.447/0.02&0.447/0.02&0.409/0.02&0.396/0.02&0.452/0.02&0.414/0.02\\ 
$p=1.1$&0.635/0.02&0.645/0.02&0.645/0.02&0.620/0.02&0.633/0.02&0.610/0.02&0.483/0.02\\ 
$p=1.2$&0.853/0.01&0.870/0.01&0.868/0.01&0.808/0.02&0.877/0.01&0.700/0.02&0.859/0.02\\ 
$p=1.3$&0.862/0.01&0.885/0.01&0.882/0.01&0.809/0.02&0.892/0.01&0.691/0.02&0.941/0.01\\ 
$p=1.4$&0.874/0.01&0.897/0.01&0.890/0.01&0.812/0.02&0.866/0.02&0.686/0.02&\textbf{0.950}/0.01\\ 
$p=1.5$&0.875/0.01&0.895/0.01&0.887/0.01&0.800/0.02&0.853/0.02&0.707/0.02&0.940/0.01\\ 
$p=1.6$&0.857/0.02&0.882/0.01&0.872/0.01&0.818/0.02&0.860/0.02&0.693/0.02&0.932/0.01\\ 
$p=1.7$&0.863/0.02&0.886/0.01&0.868/0.02&0.802/0.02&0.826/0.02&0.689/0.02&0.915/0.01\\ 
$p=1.8$&0.856/0.02&0.879/0.01&0.861/0.02&0.804/0.02&0.827/0.02&0.681/0.02&0.920/0.01\\ 
$p=1.9$&0.843/0.02&0.866/0.02&0.843/0.02&0.801/0.02&0.805/0.02&0.687/0.02&0.903/0.01\\ 
$p=2$&0.838/0.02&0.866/0.01&0.838/0.02&0.793/0.02&0.793/0.02&0.687/0.02&0.892/0.01\\ 
$p=2.5$&0.784/0.02&0.831/0.02&0.763/0.02&0.762/0.02&0.731/0.02&0.645/0.02&0.846/0.01\\ 
$p=3$&0.754/0.02&0.790/0.02&0.714/0.02&0.750/0.02&0.677/0.02&0.618/0.02&0.815/0.01\\ 
\cline{1-8}
\multicolumn{6}{l}{iMWK-Means with explicit re-scaling}\\
$p\rightarrow 1$&0.305/0.02&0.296/0.02&0.296/0.02&0.288/0.02&0.281/0.02&0.271/0.02&0.246/0.01\\ 
$p=1.1$&0.501/0.02&0.384/0.02&0.392/0.02&0.624/0.02&0.620/0.02&0.327/0.01&0.344/0.02\\ 
$p=1.2$&0.805/0.02&0.677/0.02&0.648/0.02&0.916/0.01&0.869/0.01&0.569/0.02&0.615/0.02\\ 
$p=1.3$&0.898/0.01&0.873/0.01&0.868/0.01&0.919/0.01&0.896/0.01&0.825/0.02&0.892/0.01\\ 
$p=1.4$&0.919/0.01&0.907/0.01&0.911/0.01&0.916/0.01&0.924/0.01&0.872/0.01&0.875/0.02\\ 
$p=1.5$&0.933/0.01&0.924/0.01&0.925/0.01&\textbf{0.925}/0.01&0.922/0.01&0.884/0.01&0.867/0.01\\ 
$p=1.6$&0.928/0.01&0.927/0.01&0.930/0.01&0.923/0.01&0.928/0.01&0.868/0.02&0.851/0.02\\ 
$p=1.7$&\textbf{0.934}/0.01&\textbf{0.928}/0.01&\textbf{0.931}/0.01&0.909/0.01&0.910/0.01&0.864/0.02&0.870/0.01\\ 
$p=1.8$&0.930/0.01&0.927/0.01&0.924/0.01&0.904/0.01&0.906/0.01&0.860/0.02&0.862/0.01\\ 
$p=1.9$&0.926/0.01&0.922/0.01&0.924/0.01&0.904/0.01&0.898/0.01&0.852/0.02&0.850/0.01\\ 
$p=2$&0.922/0.01&0.916/0.01&0.922/0.01&0.907/0.01&0.907/0.01&0.850/0.02&0.840/0.01\\ 
$p=2.5$&0.889/0.01&0.884/0.01&0.884/0.01&0.861/0.02&0.847/0.02&0.791/0.02&0.810/0.02\\ 
$p=3$&0.845/0.02&0.831/0.02&0.837/0.02&0.818/0.02&0.814/0.02&0.747/0.02&0.776/0.02\\ 
\cline{1-8}
\multicolumn{8}{l}{iMWK-Means with explicit re-scaling followed by K-Means}\\
$p\rightarrow 1$&0.262/0.02&0.285/0.02&0.267/0.02&0.226/0.02&0.228/0.02&0.216/0.01&0.225/0.01\\ 
$p=1.1$&0.358/0.02&0.418/0.02&0.412/0.02&0.575/0.02&0.584/0.02&0.176/0.01&0.281/0.01\\ 
$p=1.2$&0.787/0.02&0.842/0.02&0.845/0.02&0.908/0.01&0.896/0.01&0.287/0.02&0.538/0.02\\ 
$p=1.3$&0.905/0.01&0.902/0.01&0.905/0.01&0.904/0.01&0.902/0.01&0.666/0.03&0.690/0.02\\ 
$p=1.4$&0.923/0.01&0.916/0.01&0.923/0.01&0.922/0.01&\textbf{0.936}/0.01&0.857/0.02&0.599/0.02\\ 
$p=1.5$&0.923/0.01&0.918/0.01&0.922/0.01&0.919/0.01&0.927/0.01&\textbf{0.886}/0.01&0.540/0.02\\ 
$p=1.6$&0.915/0.01&0.912/0.01&0.916/0.01&0.924/0.01&0.922/0.01&0.873/0.02&0.513/0.01\\ 
$p=1.7$&0.919/0.01&0.922/0.01&0.917/0.01&0.905/0.01&0.905/0.01&0.867/0.02&0.484/0.01\\ 
$p=1.8$&0.925/0.01&0.919/0.01&0.919/0.01&0.918/0.01&0.915/0.01&0.862/0.02&0.461/0.01\\ 
$p=1.9$&0.916/0.01&0.912/0.01&0.914/0.01&0.914/0.01&0.911/0.01&0.852/0.02&0.462/0.01\\ 
$p=2$&0.909/0.01&0.902/0.01&0.909/0.01&0.900/0.01&0.900/0.01&0.856/0.02&0.453/0.01\\ 
$p=2.5$&0.894/0.01&0.914/0.01&0.885/0.01&0.852/0.02&0.836/0.02&0.802/0.02&0.432/0.01\\ 
$p=3$&0.879/0.02&0.920/0.01&0.849/0.02&0.807/0.02&0.804/0.02&0.770/0.02&0.382/0.01\\
\end{tabular}
\end{center}
\label{Tab:ARILowNoise}
\end{table}
%
%
%

\begin{table}[h]\footnotesize
\caption{The percentage of finding the true number of clusters in data sets with about 50\% extra features composed of uniformly random noise.}
\begin{center}
\tabcolsep=0.04cm 
\begin{tabular}{@{\extracolsep{2pt}}lccccccc@{}}
&\multicolumn{3}{c}{Silhouette}&\multicolumn{2}{c}{Dunn}\\
\cline{2-4}
\cline{5-6}
&Eucl&Manh&Mink&Eucl&Mink&CH&Hartigan\\
\cline{2-8}
K-Means&65.500&79.500&65.500&33.500&33.500&29.500&4.000\\
\cline{1-8}
\multicolumn{6}{l}{iMWK-Means}\\
$p\rightarrow 1$&28.500&28.500&28.500&30.500&23.500&28.500&34.500\\ 
$p=1.1$&30.000&31.500&31.500&32.000&34.500&28.500&32.000\\ 
$p=1.2$&67.500&72.000&71.500&56.000&69.000&38.000&70.000\\ 
$p=1.3$&68.000&74.500&73.500&55.500&72.000&36.500&\textbf{81.500}\\ 
$p=1.4$&71.500&76.500&75.000&57.500&68.500&37.500&80.500\\ 
$p=1.5$&70.000&75.500&73.000&54.500&64.500&38.500&74.000\\ 
$p=1.6$&68.500&73.500&71.000&56.500&64.500&38.500&71.500\\ 
$p=1.7$&70.000&74.000&71.000&53.500&59.000&38.500&64.500\\ 
$p=1.8$&67.500&73.000&68.500&53.500&58.500&37.500&65.000\\ 
$p=1.9$&65.500&69.000&65.500&51.500&53.000&39.500&60.500\\ 
$p=2$&62.500&66.000&62.500&48.000&48.000&38.500&56.500\\ 
$p=2.5$&58.500&61.000&55.500&44.000&40.000&36.000&49.500\\ 
$p=3$&54.500&56.000&51.000&40.500&32.500&33.500&40.500\\ 
\cline{1-8}
\multicolumn{8}{l}{iMWK-Means with explicit re-scaling}\\
$p\rightarrow 1$&19.000&14.000&14.000&14.000&12.500&14.500&11.500\\ 
$p=1.1$&29.000&17.500&19.000&32.500&32.000&10.000&9.000\\ 
$p=1.2$&59.500&34.000&39.500&74.000&61.500&28.500&20.500\\ 
$p=1.3$&71.500&42.500&56.000&66.000&54.000&63.500&74.500\\ 
$p=1.4$&76.500&46.000&60.500&59.000&55.500&71.500&38.500\\ 
$p=1.5$&73.000&51.500&62.500&60.500&58.000&74.000&0.500\\ 
$p=1.6$&71.000&54.500&64.500&59.500&52.500&72.500&0.500\\ 
$p=1.7$&68.000&55.500&63.500&47.500&43.500&72.000&0.500\\ 
$p=1.8$&65.500&58.500&62.000&46.000&41.500&70.000&0.500\\ 
$p=1.9$&59.000&58.500&58.000&39.500&37.000&67.000&0.500\\ 
$p=2$&61.000&61.000&61.000&39.500&39.500&64.500&1.000\\ 
$p=2.5$&57.000&59.000&57.000&31.000&33.500&58.000&0.500\\ 
$p=3$&52.500&54.500&46.000&27.000&29.000&52.000&0.000\\ 
\cline{1-8}
\multicolumn{8}{l}{iMWK-Means with explicit re-scaling followed by K-Means}\\
$p\rightarrow 1$&16.000&14.500&13.500&10.500&11.500&8.500&10.500\\ 
$p=1.1$&24.000&22.500&24.000&33.500&34.000&1.000&9.000\\ 
$p=1.2$&60.000&64.000&64.500&80.000&77.500&9.000&27.000\\ 
$p=1.3$&76.500&76.500&76.500&79.500&78.500&48.000&27.000\\ 
$p=1.4$&\textbf{82.500}&81.500&\textbf{83.000}&\textbf{82.500}&\textbf{85.500}&70.000&8.500\\ 
$p=1.5$&80.500&80.500&80.500&81.000&83.500&\textbf{74.500}&6.500\\ 
$p=1.6$&80.000&80.000&80.000&81.000&80.500&73.500&3.500\\ 
$p=1.7$&81.500&81.500&81.500&76.000&75.500&73.000&3.000\\ 
$p=1.8$&80.500&80.500&79.000&79.500&79.000&71.000&1.500\\ 
$p=1.9$&77.500&78.500&77.000&78.500&77.500&68.500&0.500\\ 
$p=2$&76.000&76.500&76.000&76.000&76.000&68.500&1.000\\ 
$p=2.5$&75.000&80.500&74.000&69.000&65.500&63.000&1.500\\ 
$p=3$&75.000&\textbf{83.000}&69.000&61.500&59.500&59.000&1.000\\ 
\end{tabular}
\end{center}
\label{Tab:PercentFindRightClusters_LowNoise}
\end{table}

In our third set of experiments, we have added 100\% extra noise features to each data set, effectively doubling their number of dimensions. Tables \ref{Tab:RelativeErrorHighNoise}, \ref{Tab:ARIHighNoise}, and \ref{Tab:PercentFindRightClusters_HighNoise} show the results for this set of experiments regarding the relative error, the adjusted Rand index, and the percentage of times the algorithm found the true number of clusters.

The superiority of Silhouette (Manhattan) among the baseline K-Means methods is even stronger here than with 50\% noise, and it is actually quite competitive, although it cannot beat the best iMWK-based methods. 

Table \ref{Tab:RelativeErrorHighNoise} shows that iMWK-Means with explicit re-scaling followed by K-Means is still the method that produces the lowest relative errors for all cluster validity indexes, except for Hartigan. In this experiment the value of $p$ that produces most optimal results per index is $p=1.7$, although the best overall result is achieved at $p=3.0$ using the Silhouette and the Manhattan distance. The entries under $p=1.4$ still produce results that are better than the K-Means baseline for all cluster validity indexes except Silhouette (Manhattan), and values between $1.5$ and $1.7$ usually do better. In fact, for the Silhouette index using the Euclidean distance, any $p$ higher than $1.2$ produces relative errors at least $4.4$ times smaller than the baseline, and a much smaller standard error.

Table \ref{Tab:ARIHighNoise} shows that as in our previous experiments, the best adjusted Rand index is given by the iMWK-Means with explicit re-scaling without final K-Means with best values occurring at $p=1.5$ for the Silhouette index (all versions). However, just like before, the adjusted Rand indexes for iMWK-Means with explicit re-scaling followed by K-Means are considerably better than the K-Means baseline for most values of $p$, with best results at $p=1.7$ for the Silhouette index. 

Table \ref{Tab:PercentFindRightClusters_HighNoise} on the percentage of finding the true number of clusters also shows that all cluster validity indexes benefit from iMWK-Means with explicit feature re-scaling followed by K-Means, at various values of $p$. The best result overall was given by the Silhouette index using the Manhattan distance at $p=3$, as was also the best value of $p$ for this index in the experiments with 50\% noise features. Most of the other indexes achieved their best results at $p=1.8$ or $p=2$, slightly higher than what was best for the relative error.

\begin{table}[h]\footnotesize
\caption{Relative errors of the estimation of the number of clusters and their standard errors in GM data with extra 100\% noise features, in relation the the known number of classes.}
\begin{center}
\tabcolsep=0.04cm 
\begin{tabular}{@{\extracolsep{2pt}}lccccccc@{}}
&\multicolumn{3}{c}{Silhouette}&\multicolumn{2}{c}{Dunn}\\
\cline{2-4}
\cline{5-6}
&Eucl&Manh&Mink&Eucl&Mink&CH&Hartigan\\
\cline{2-8}
K-Means&0.671/0.15&0.086/0.01&0.671/0.15&2.800/0.21&2.800/0.21&0.347/0.02&3.002/0.17\\ 
\cline{1-8}
\multicolumn{6}{l}{iMWK-Means}\\
$p\rightarrow 1$&0.332/0.02&0.326/0.02&0.326/0.02&0.664/0.05&0.707/0.06&0.315/0.02&0.380/0.02\\ 
$p=1.1$&0.284/0.02&0.280/0.02&0.280/0.02&0.381/0.04&0.358/0.03&0.308/0.02&0.321/0.02\\ 
$p=1.2$&0.190/0.02&0.170/0.01&0.168/0.01&0.255/0.02&0.247/0.03&0.285/0.02&0.143/0.01\\ 
$p=1.3$&0.187/0.01&0.174/0.01&0.175/0.01&0.413/0.05&0.345/0.04&0.281/0.02&\textbf{0.130}/0.02\\ 
$p=1.4$&0.210/0.02&0.189/0.01&0.198/0.01&0.432/0.05&0.412/0.04&0.297/0.02&0.143/0.01\\ 
$p=1.5$&0.197/0.01&0.181/0.01&0.183/0.01&0.399/0.04&0.443/0.05&0.303/0.02&0.131/0.01\\ 
$p=1.6$&0.194/0.01&0.183/0.01&0.186/0.01&0.399/0.04&0.464/0.05&0.294/0.02&0.150/0.02\\ 
$p=1.7$&0.177/0.01&0.172/0.01&0.175/0.01&0.487/0.07&0.514/0.07&0.299/0.02&0.179/0.02\\ 
$p=1.8$&0.184/0.01&0.169/0.01&0.176/0.01&0.544/0.07&0.542/0.07&0.301/0.02&0.209/0.03\\ 
$p=1.9$&0.183/0.01&0.160/0.01&0.179/0.01&0.527/0.07&0.548/0.07&0.297/0.02&0.234/0.03\\ 
$p=2$&0.204/0.01&0.181/0.01&0.204/0.01&0.601/0.07&0.601/0.07&0.308/0.02&0.285/0.03\\ 
$p=2.5$&0.210/0.01&0.216/0.02&0.223/0.01&0.781/0.09&0.807/0.09&0.322/0.02&0.420/0.05\\ 
$p=3$&0.258/0.02&0.277/0.02&0.263/0.02&0.982/0.11&0.886/0.09&0.318/0.02&0.462/0.04\\ 
\cline{1-8}
\multicolumn{6}{l}{iMWK-Means with explicit re-scaling}\\
$p\rightarrow 1$&1.754/0.11&1.700/0.11&1.700/0.11&1.619/0.10&1.644/0.10&1.824/0.10&1.493/0.09\\ 
$p=1.1$&1.727/0.12&2.485/0.12&2.431/0.12&0.299/0.02&0.281/0.02&1.228/0.07&1.526/0.08\\ 
$p=1.2$&0.782/0.09&1.852/0.13&1.828/0.14&0.185/0.02&0.211/0.02&0.882/0.10&1.608/0.07\\ 
$p=1.3$&0.663/0.09&1.850/0.10&1.793/0.13&0.324/0.03&0.462/0.06&0.259/0.06&0.166/0.02\\ 
$p=1.4$&0.906/0.11&2.462/0.11&1.903/0.11&0.394/0.05&0.618/0.08&0.170/0.01&0.575/0.07\\ 
$p=1.5$&0.908/0.08&2.587/0.11&1.877/0.11&0.439/0.05&0.576/0.06&0.164/0.02&1.123/0.08\\ 
$p=1.6$&1.225/0.10&2.634/0.12&1.911/0.11&0.459/0.06&0.761/0.09&0.156/0.01&1.517/0.07\\ 
$p=1.7$&1.462/0.11&2.537/0.12&1.811/0.11&0.640/0.08&0.872/0.10&0.165/0.01&1.858/0.08\\ 
$p=1.8$&1.616/0.12&2.540/0.13&1.745/0.12&0.982/0.11&1.164/0.12&0.173/0.02&2.167/0.08\\ 
$p=1.9$&1.661/0.12&2.129/0.13&1.683/0.12&1.083/0.11&1.212/0.12&0.172/0.02&2.311/0.08\\ 
$p=2$&1.611/0.11&2.078/0.12&1.611/0.11&1.363/0.13&1.363/0.13&0.188/0.01&2.561/0.09\\ 
$p=2.5$&1.873/0.11&1.476/0.11&1.800/0.11&2.106/0.14&1.707/0.14&0.239/0.02&2.913/0.09\\ 
$p=3$&1.209/0.10&0.580/0.07&1.581/0.10&2.522/0.16&1.675/0.15&0.293/0.02&2.881/0.10\\ 
\cline{1-8}
\multicolumn{8}{l}{iMWK-Means with explicit re-scaling followed by K-Means}\\
$p\rightarrow 1$&1.942/0.11&1.436/0.09&1.832/0.11&1.973/0.10&1.865/0.10&2.142/0.10&1.687/0.08\\ 
$p=1.1$&1.754/0.10&1.550/0.12&1.905/0.12&0.295/0.02&0.284/0.02&3.016/0.11&1.602/0.09\\ 
$p=1.2$&0.693/0.11&0.171/0.03&0.296/0.07&0.144/0.01&0.144/0.01&3.300/0.12&1.081/0.10\\ 
$p=1.3$&0.121/0.01&0.111/0.01&0.114/0.01&0.156/0.02&0.148/0.01&1.444/0.16&1.340/0.11\\ 
$p=1.4$&0.118/0.01&0.132/0.03&0.108/0.01&0.150/0.02&0.140/0.01&0.394/0.10&1.755/0.10\\ 
$p=1.5$&0.101/0.01&0.090/0.01&0.103/0.01&0.123/0.01&0.110/0.01&0.130/0.01&2.113/0.11\\ 
$p=1.6$&0.108/0.01&0.100/0.01&0.101/0.01&0.181/0.04&0.152/0.04&0.127/0.01&2.517/0.10\\ 
$p=1.7$&\textbf{0.091}/0.01&0.081/0.01&\textbf{0.087}/0.01&\textbf{0.109}/0.01&\textbf{0.108}/0.01&0.139/0.01&2.712/0.10\\ 
$p=1.8$&0.094/0.01&0.077/0.01&0.097/0.01&0.125/0.02&0.121/0.02&\textbf{0.122}/0.01&2.874/0.12\\ 
$p=1.9$&0.120/0.02&0.084/0.01&0.122/0.02&0.149/0.03&0.150/0.03&0.130/0.01&3.026/0.11\\ 
$p=2$&0.126/0.02&0.074/0.01&0.126/0.02&0.134/0.02&0.134/0.02&0.146/0.01&3.081/0.11\\ 
$p=2.5$&0.146/0.02&0.071/0.01&0.154/0.02&0.575/0.11&0.551/0.11&0.191/0.02&3.507/0.11\\ 
$p=3$&0.151/0.03&\textbf{0.068}/0.01&0.168/0.02&0.730/0.13&0.546/0.09&0.218/0.02&3.721/0.13\\  
\end{tabular}
\end{center}
\label{Tab:RelativeErrorHighNoise}
\end{table}

\begin{table}[h]\footnotesize
\caption{The adjusted Rand index and its standard error for the clusterings at each estimated $K$ in the data sets with about 100\% extra features composed of uniformly random noise.}
\begin{center}
\tabcolsep=0.04cm 
\begin{tabular}{@{\extracolsep{2pt}}lccccccc@{}}
&\multicolumn{3}{c}{Silhouette}&\multicolumn{2}{c}{Dunn}\\
\cline{2-4}
\cline{5-6}
&Eucl&Manh&Mink&Eucl&Mink&CH&Hartigan\\
\cline{2-8}
K-Means&0.771/0.02&0.867/0.01&0.771/0.02&0.445/0.02&0.445/0.02&0.527/0.02&0.389/0.02\\ 
\cline{1-8}
\multicolumn{6}{l}{iMWK-Means}\\
$p\rightarrow 1$&0.402/0.02&0.407/0.02&0.407/0.02&0.360/0.02&0.349/0.02&0.416/0.02&0.367/0.02\\ 
$p=1.1$&0.631/0.02&0.635/0.02&0.635/0.02&0.616/0.02&0.625/0.02&0.607/0.02&0.458/0.02\\ 
$p=1.2$&0.766/0.02&0.800/0.02&0.793/0.02&0.778/0.02&0.822/0.02&0.653/0.02&0.825/0.02\\ 
$p=1.3$&0.766/0.02&0.786/0.02&0.783/0.02&0.800/0.02&0.876/0.01&0.652/0.02&0.850/0.01\\ 
$p=1.4$&0.733/0.02&0.764/0.02&0.751/0.02&0.791/0.02&0.854/0.02&0.628/0.02&0.851/0.01\\ 
$p=1.5$&0.756/0.02&0.775/0.02&0.773/0.02&0.779/0.02&0.826/0.02&0.633/0.02&0.864/0.01\\ 
$p=1.6$&0.745/0.02&0.771/0.02&0.761/0.02&0.780/0.02&0.810/0.02&0.630/0.02&0.856/0.01\\ 
$p=1.7$&0.774/0.02&0.807/0.01&0.791/0.02&0.777/0.02&0.798/0.02&0.627/0.02&0.862/0.01\\ 
$p=1.8$&0.759/0.02&0.783/0.02&0.769/0.02&0.763/0.02&0.784/0.02&0.622/0.02&0.854/0.01\\ 
$p=1.9$&0.751/0.02&0.785/0.02&0.757/0.02&0.754/0.02&0.756/0.02&0.611/0.02&0.851/0.01\\ 
$p=2$&0.713/0.02&0.747/0.02&0.713/0.02&0.741/0.02&0.741/0.02&0.586/0.02&0.835/0.01\\ 
$p=2.5$&0.626/0.02&0.697/0.02&0.599/0.02&0.696/0.02&0.675/0.02&0.514/0.02&0.794/0.02\\ 
$p=3$&0.590/0.02&0.635/0.02&0.548/0.02&0.660/0.02&0.615/0.02&0.512/0.02&0.724/0.02\\ 
\cline{1-8}
\multicolumn{6}{l}{iMWK-Means with explicit re-scaling}\\
$p\rightarrow 1$&0.268/0.02&0.258/0.02&0.258/0.02&0.234/0.02&0.234/0.02&0.247/0.02&0.195/0.01\\ 
$p=1.1$&0.422/0.02&0.299/0.02&0.302/0.02&0.628/0.02&0.635/0.02&0.322/0.02&0.289/0.01\\ 
$p=1.2$&0.757/0.02&0.632/0.02&0.622/0.02&\textbf{0.891}/0.01&0.863/0.01&0.679/0.02&0.484/0.02\\ 
$p=1.3$&0.859/0.02&0.846/0.01&0.829/0.02&0.887/0.01&0.862/0.01&0.827/0.02&0.866/0.01\\ 
$p=1.4$&0.896/0.01&0.912/0.01&0.919/0.01&0.878/0.01&0.877/0.01&0.821/0.02&0.878/0.01\\ 
$p=1.5$&\textbf{0.934}/0.01&\textbf{0.924}/0.01&\textbf{0.934}/0.01&0.883/0.01&\textbf{0.892}/0.01&0.807/0.02&0.877/0.01\\ 
$p=1.6$&0.921/0.01&0.916/0.01&0.915/0.01&0.869/0.01&0.884/0.01&0.815/0.02&0.880/0.01\\ 
$p=1.7$&0.914/0.01&0.915/0.01&0.916/0.01&0.886/0.01&0.879/0.01&0.828/0.01&\textbf{0.882}/0.01\\ 
$p=1.8$&0.902/0.01&0.901/0.01&0.901/0.01&0.847/0.01&0.852/0.01&0.785/0.02&0.868/0.01\\ 
$p=1.9$&0.902/0.01&0.898/0.01&0.902/0.01&0.859/0.01&0.857/0.01&0.781/0.02&0.852/0.01\\ 
$p=2$&0.891/0.01&0.890/0.01&0.891/0.01&0.845/0.01&0.845/0.01&0.755/0.02&0.854/0.01\\ 
$p=2.5$&0.823/0.01&0.816/0.01&0.810/0.02&0.799/0.02&0.778/0.02&0.658/0.02&0.799/0.01\\ 
$p=3$&0.727/0.02&0.693/0.02&0.739/0.02&0.712/0.02&0.699/0.02&0.607/0.02&0.731/0.02\\ 
\cline{1-8}
\multicolumn{8}{l}{iMWK-Means with explicit re-scaling followed by K-Means}\\
$p\rightarrow 1$&0.231/0.02&0.251/0.02&0.230/0.02&0.178/0.02&0.188/0.02&0.188/0.02&0.143/0.01\\ 
$p=1.1$&0.297/0.02&0.365/0.02&0.333/0.03&0.599/0.02&0.602/0.02&0.127/0.01&0.220/0.01\\ 
$p=1.2$&0.742/0.02&0.830/0.02&0.822/0.02&0.833/0.02&0.825/0.02&0.212/0.02&0.617/0.03\\ 
$p=1.3$&0.852/0.02&0.856/0.01&0.850/0.02&0.813/0.02&0.824/0.02&0.660/0.03&0.635/0.02\\ 
$p=1.4$&0.870/0.01&0.869/0.01&0.876/0.01&0.834/0.02&0.841/0.02&0.822/0.02&0.531/0.02\\ 
$p=1.5$&0.875/0.01&0.878/0.01&0.871/0.01&0.857/0.01&0.869/0.01&0.845/0.02&0.467/0.01\\ 
$p=1.6$&0.879/0.01&0.877/0.01&0.881/0.01&0.844/0.02&0.866/0.01&0.843/0.02&0.396/0.01\\ 
$p=1.7$&0.893/0.01&0.897/0.01&0.895/0.01&0.875/0.01&0.880/0.01&\textbf{0.847}/0.01&0.381/0.01\\ 
$p=1.8$&0.890/0.01&0.895/0.01&0.886/0.01&0.877/0.01&0.881/0.01&0.838/0.02&0.380/0.01\\ 
$p=1.9$&0.882/0.01&0.894/0.01&0.881/0.01&0.875/0.01&0.874/0.01&0.825/0.02&0.361/0.01\\ 
$p=2$&0.874/0.01&0.897/0.01&0.874/0.01&0.875/0.01&0.875/0.01&0.796/0.02&0.357/0.01\\ 
$p=2.5$&0.832/0.02&0.894/0.01&0.811/0.02&0.796/0.02&0.795/0.02&0.724/0.02&0.323/0.01\\ 
$p=3$&0.821/0.02&0.893/0.01&0.779/0.02&0.765/0.02&0.737/0.02&0.695/0.02&0.315/0.01\\ 
\end{tabular}
\end{center}
\label{Tab:ARIHighNoise}
\end{table}
%
%
%

\begin{table}[h]\footnotesize
\caption{The percentage of finding the true number of clusters in data sets with about 100\% extra features composed of uniformly random noise.}
\begin{center}
\tabcolsep=0.04cm 
\begin{tabular}{@{\extracolsep{2pt}}lccccccc@{}}
&\multicolumn{3}{c}{Silhouette}&\multicolumn{2}{c}{Dunn}\\
\cline{2-4}
\cline{5-6}
&Eucl&Manh&Mink&Eucl&Mink&CH&Hartigan\\
\cline{2-8}
K-Means&55.000&74.500&55.000&21.000&21.000&25.000&1.500\\ 
\cline{1-8}
\multicolumn{6}{l}{iMWK-Means}\\
$p\rightarrow 1$&28.000&28.000&28.000&21.500&25.500&27.500&21.500\\ 
$p=1.1$&32.500&33.000&33.000&31.500&33.500&29.500&28.500\\ 
$p=1.2$&50.500&54.000&54.500&46.500&52.000&32.500&60.000\\ 
$p=1.3$&49.500&52.000&51.500&43.500&51.000&34.000&\textbf{62.000}\\ 
$p=1.4$&46.000&48.500&47.000&40.500&45.500&30.000&57.000\\ 
$p=1.5$&48.000&50.500&50.000&43.000&45.500&30.500&61.500\\ 
$p=1.6$&46.000&48.500&47.500&39.000&41.500&30.500&57.500\\ 
$p=1.7$&47.500&49.500&48.500&38.000&39.500&31.000&52.500\\ 
$p=1.8$&48.500&52.500&50.500&39.000&41.500&31.500&53.000\\ 
$p=1.9$&48.500&53.500&49.500&40.500&40.500&31.500&53.000\\ 
$p=2$&44.000&49.000&44.000&34.500&34.500&30.000&46.500\\ 
$p=2.5$&41.000&42.500&39.500&27.000&24.000&26.500&36.000\\ 
$p=3$&35.500&33.500&34.500&22.500&18.500&28.000&25.500\\ 
\cline{1-8}
\multicolumn{8}{l}{iMWK-Means with explicit re-scaling}\\
$p\rightarrow 1$&18.500&14.500&14.500&12.500&10.500&16.500&8.500\\ 
$p=1.1$&26.000&12.000&13.000&35.000&34.000&11.000&8.500\\ 
$p=1.2$&47.000&14.000&22.500&58.000&53.500&42.500&0.500\\ 
$p=1.3$&47.000&13.500&21.000&47.500&44.500&54.500&60.000\\ 
$p=1.4$&38.500&14.000&24.500&47.000&41.500&53.500&12.000\\ 
$p=1.5$&39.500&15.500&25.000&48.000&39.000&57.000&0.000\\ 
$p=1.6$&35.000&17.000&25.000&40.000&32.000&57.000&0.000\\ 
$p=1.7$&30.000&20.000&26.000&35.500&27.500&53.000&0.000\\ 
$p=1.8$&31.500&22.000&29.500&27.500&25.000&55.000&0.500\\ 
$p=1.9$&29.500&26.500&29.500&30.000&26.000&55.000&0.500\\ 
$p=2$&30.000&28.000&30.000&23.000&23.000&49.500&0.000\\ 
$p=2.5$&19.500&24.000&22.000&12.000&15.500&40.000&0.500\\ 
$p=3$&28.000&33.500&22.500&11.500&16.500&30.000&0.000\\ 
\cline{1-8}
\multicolumn{8}{l}{iMWK-Means with explicit re-scaling followed by K-Means}\\
$p\rightarrow 1$&14.500&13.500&12.000&8.000&7.500&9.500&2.000\\ 
$p=1.1$&18.000&22.000&21.500&36.000&37.000&1.500&5.000\\ 
$p=1.2$&56.000&60.500&61.000&64.500&63.500&6.500&38.500\\ 
$p=1.3$&63.500&64.500&63.500&61.000&61.000&42.000&16.500\\ 
$p=1.4$&63.000&68.500&66.500&61.000&62.500&58.000&4.500\\ 
$p=1.5$&66.000&70.000&65.000&65.500&67.500&65.000&1.500\\ 
$p=1.6$&67.000&70.000&70.000&63.500&69.000&65.500&0.000\\ 
$p=1.7$&69.000&73.000&70.500&68.000&68.500&61.500&0.500\\ 
$p=1.8$&\textbf{73.500}&77.000&\textbf{72.500}&68.500&69.500&\textbf{68.500}&1.000\\ 
$p=1.9$&70.000&76.000&69.500&70.000&69.000&67.000&0.500\\ 
$p=2$&68.500&76.500&68.500&\textbf{71.500}&\textbf{71.500}&63.000&0.500\\ 
$p=2.5$&65.500&77.500&63.500&57.000&58.500&51.500&0.500\\ 
$p=3$&66.500&\textbf{78.500}&58.500&50.500&44.000&47.000&0.000\\ 
\end{tabular}
\end{center}
\label{Tab:PercentFindRightClusters_HighNoise}
\end{table}

Here are some further overall results. Interestingly, whereas iMWK-Means with explicit feature re-scaling followed by another round of K-Means seemed to be much better for estimating the number of clusters across the board, results regarding the adjusted Rand index were actually clearly better without the final K-Means. iMWK-Means without explicit re-scaling was, in most cases, inferior to either one or both of the methods that involved explicit re-scaling in presence of noise, but was better than them without noise. The CH index benefitted a lot from weighted Minkowski and feature re-scaling, but did not perform that well generally. The Dunn index had occasional good results but seemed slightly inferior to the Silhouette width where both were used in the best way. 

\begin{figure}[tb]
  \centering
  \includegraphics[width = 0.49\textwidth]{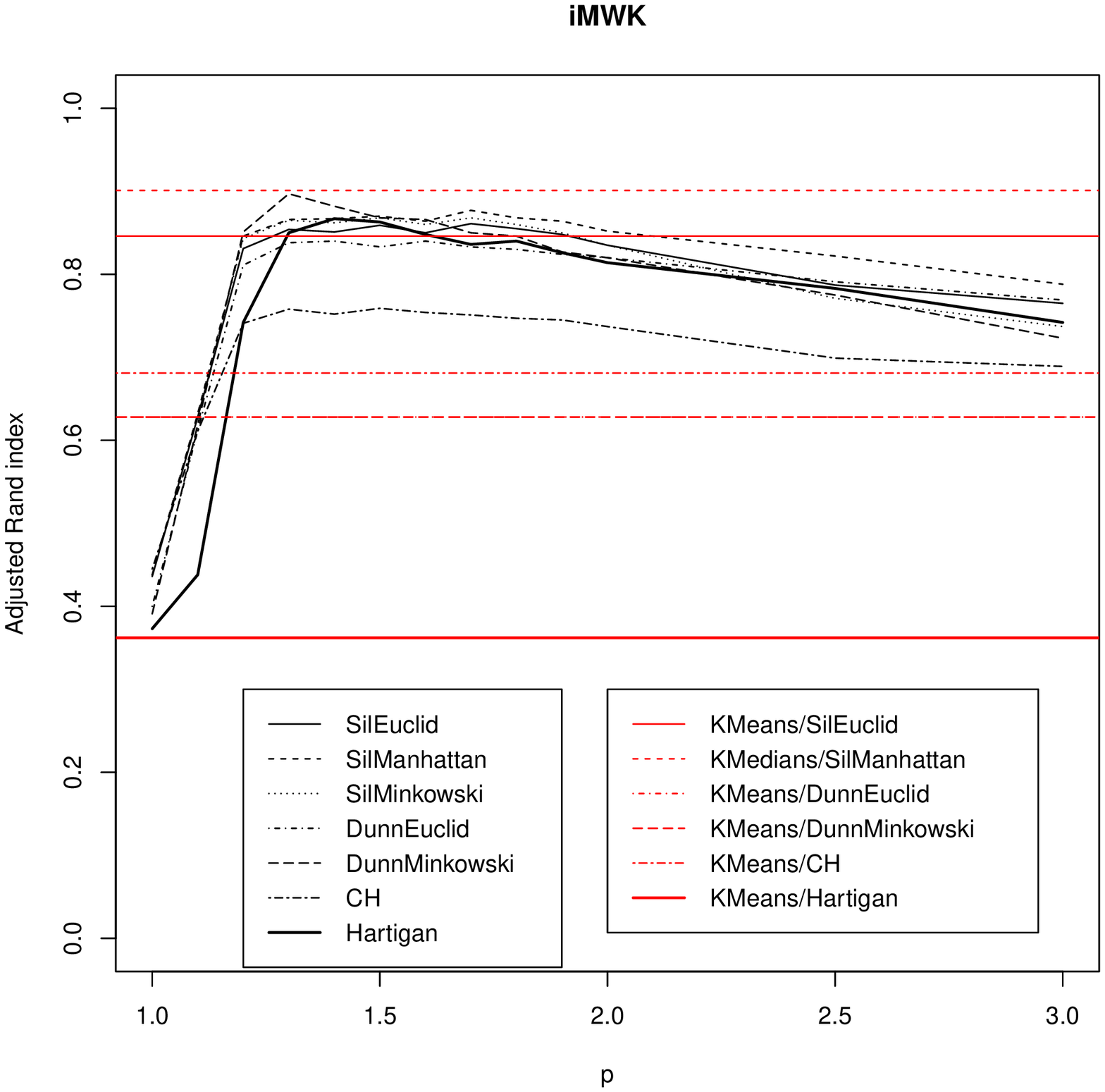}
  \includegraphics[width = 0.49\textwidth]{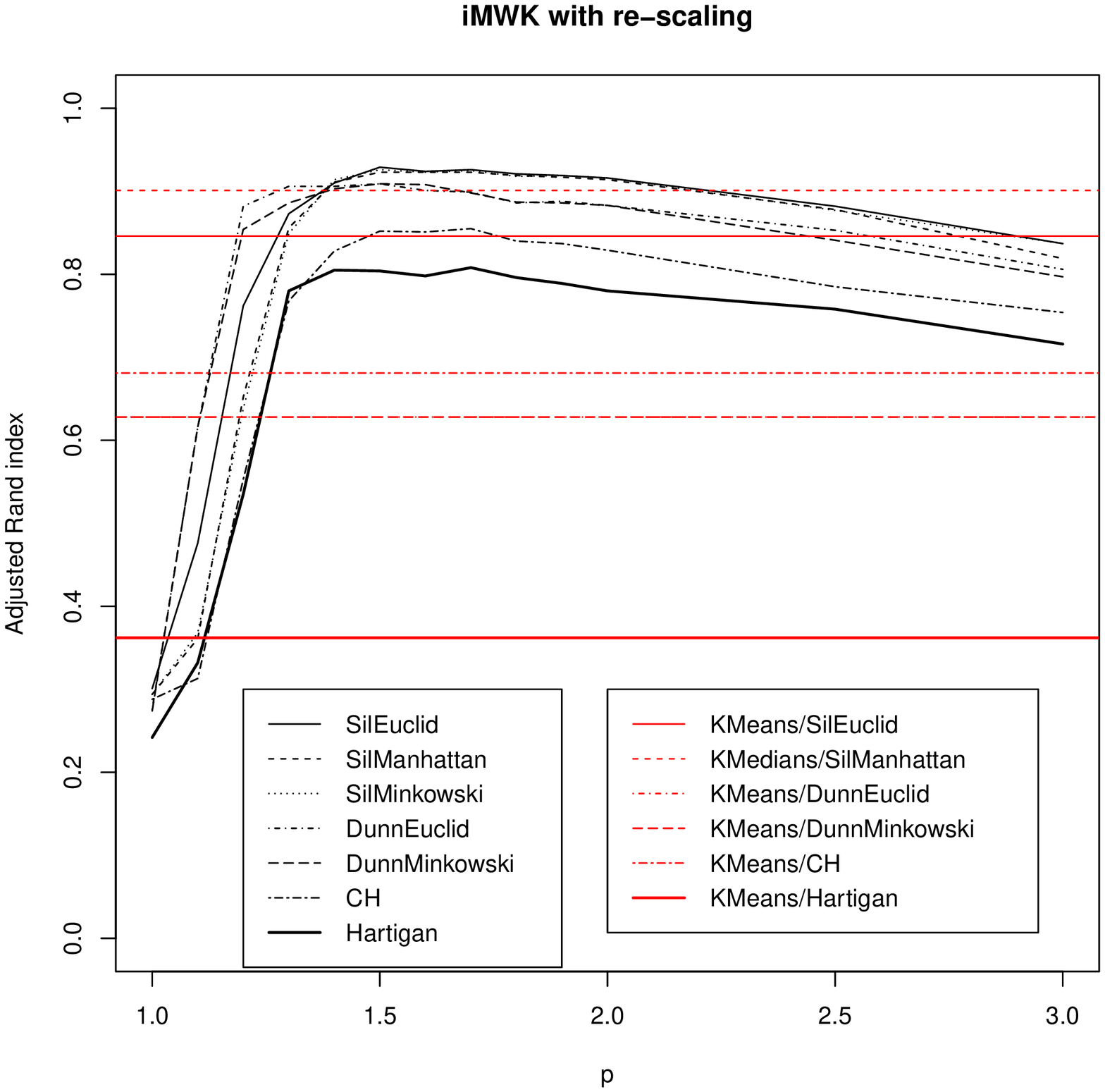}
  \includegraphics[width = 0.49\textwidth]{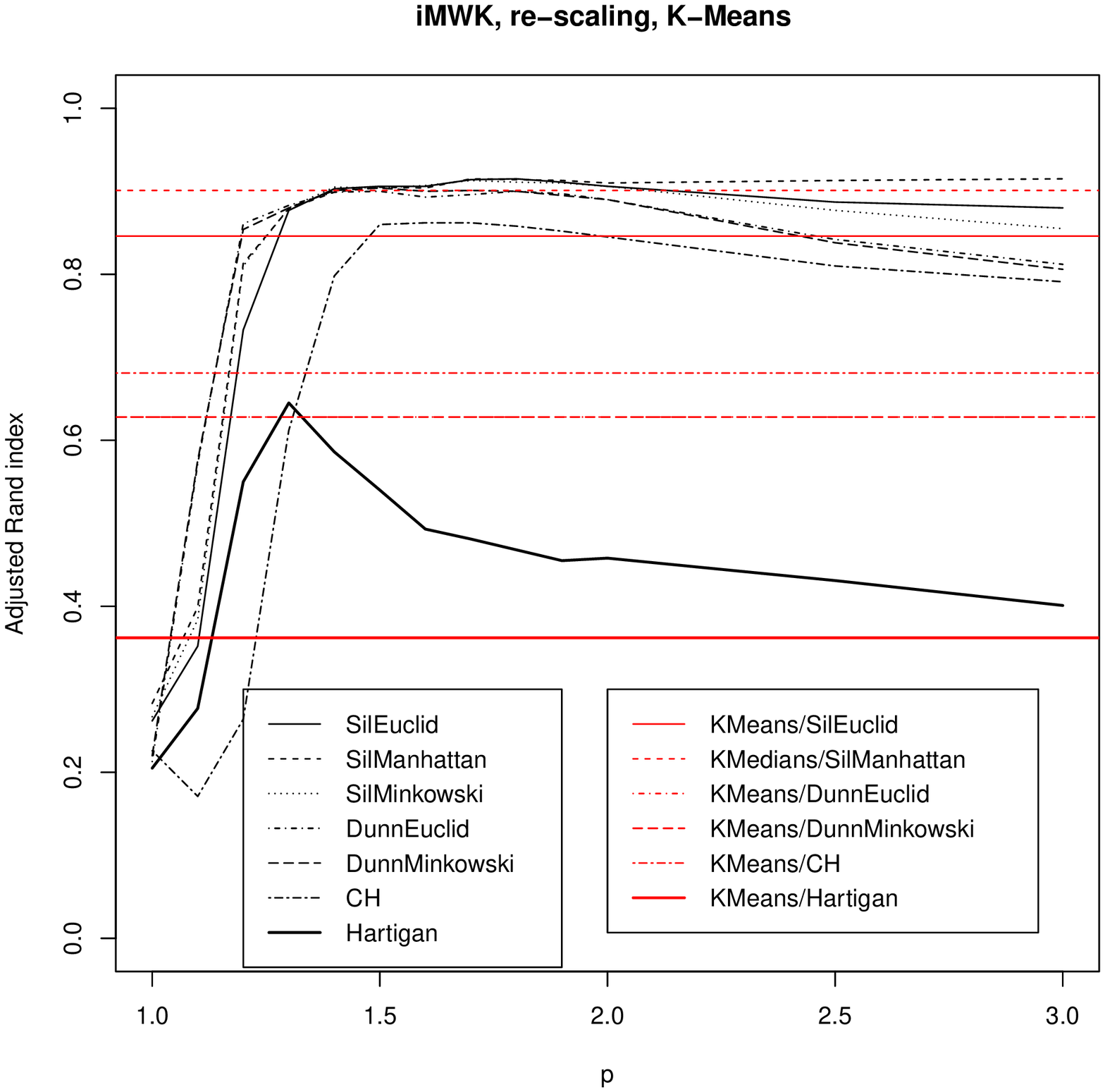}
  \caption{Adjusted Rand index values averaged over all standard scenarios.}
  \label{fig:ari}
\end{figure}

\begin{figure}[tb]
  \centering
  \includegraphics[width = 0.49\textwidth]{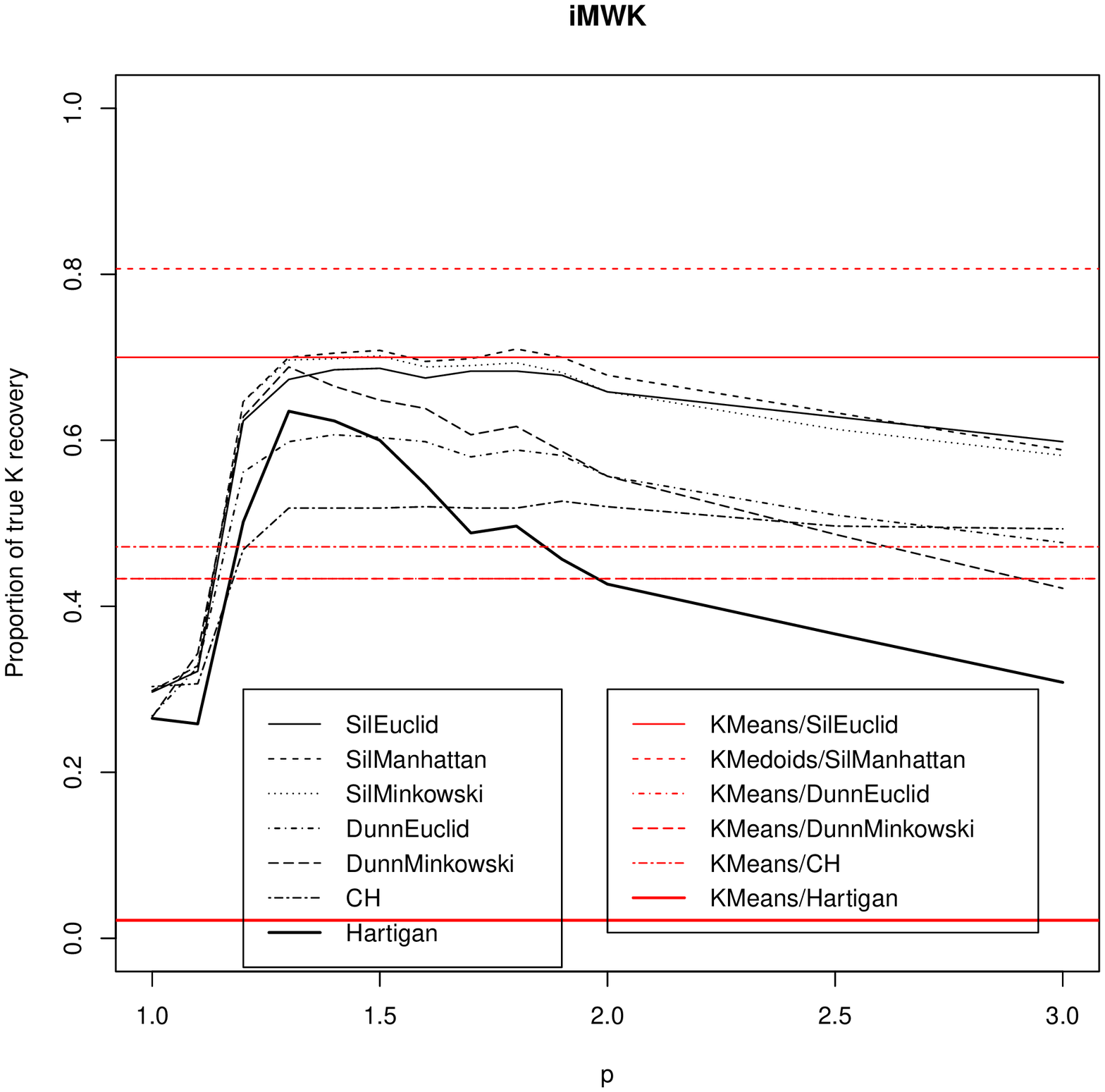}
  \includegraphics[width = 0.49\textwidth]{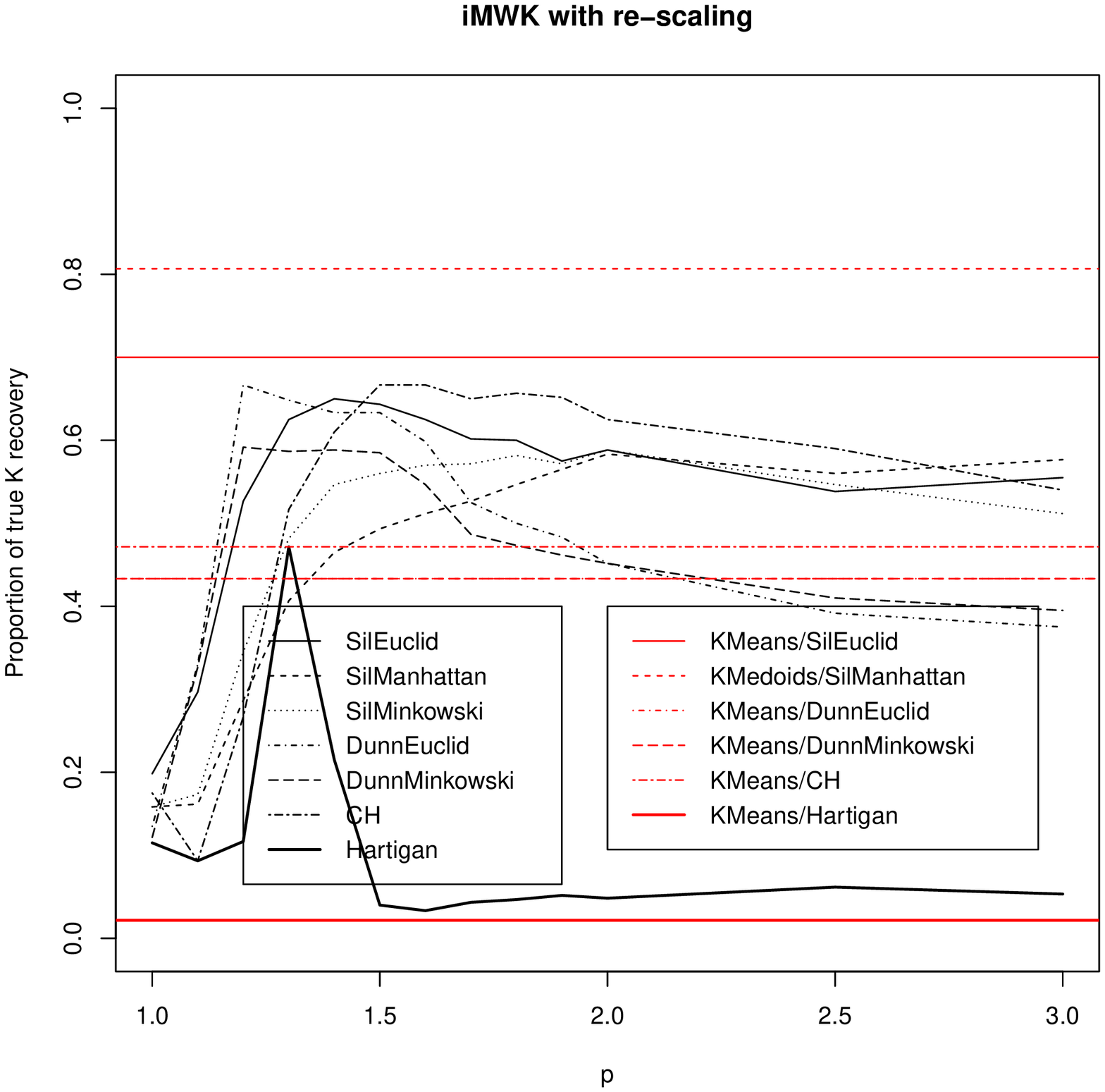}
  \includegraphics[width = 0.49\textwidth]{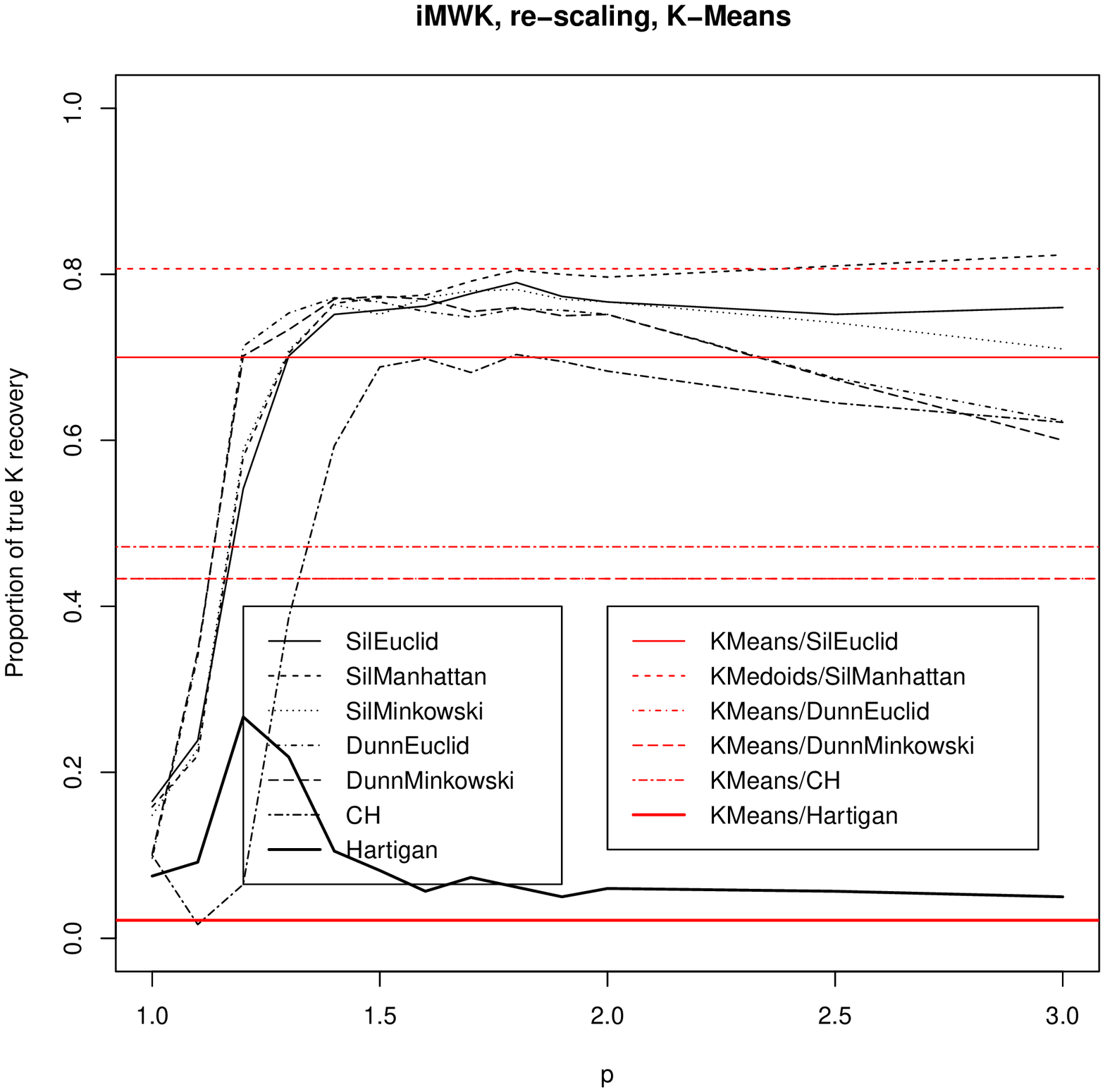}
  \caption{Percentages of finding the true number of clusters averaged over all standard scenarios.}
  \label{fig:perc}
\end{figure}

The Hartigan index behaved remarkably different. Unlike the other indexes it benefited the most from iMWK-Means without explicit re-scaling of features, and consistently showed the worst results in all of our baseline experiments. Hartigan's original threshold does not seem suitable here. Investigation of a few example datasets showed that Hartigan's index usually indeed drops at the right number of clusters, but not far enough to fulfill Hartigan's original rule, resulting in a systematic overestimation of the number of clusters. Suggesting an improved threshold value could be a target of further research. Still, it achieved the best result in one respect (adjusted Rand-index for iMWK-Means for 50\% noise 
features, Table 
\ref{Tab:ARILowNoise}) but was otherwise usually fairly weak using any reweighting scheme. Furthermore, occasional good performances regarding the percentage of finding the true number of clusters with explicit re-scaling are quite unstable toward the choice of $p$.

The results obtained with the Hartigan index seem, at first, counter-intuitive when comparing our second (50\% extra noise features) and third (100\% extra noise features) sets of experiments in terms of relative error and adjusted rand index. It looks as if adding noise features helps this index to produce better results. One should note that as the proportion of noise features increase the ratio between $W_k$ and $W_{k+1}$ decreases, which counters the overestimation of the true $K$ that we observed for this index without or with little noise.

The Figures \ref{fig:ari} and \ref{fig:perc} give a graphical summary of the results, averaged over all simulations mentioned up to this point. They show that for Dunn, CH and Hartigan, iMWK-Means and the methods introduced here are substantially better than the plain K-Means baseline over a wide range of values of $p$. As opposed to other indexes, the Silhouette (``Sil'')/Euclid does not give particularly convincing results with our methods as compared to the baseline K-Means.

K-Means together with Sil/Euclid (``K-MSE'') and particularly M-Medians with Sil/Manhattan (``K-MSM''), clearly the best of the baseline methods, raise the bar much higher, though. iMWK-Means is worse than K-MSM with all indexes and all values of $p$. For the adjusted Rand index (Figure (\ref{fig:ari}), iMWK-Means with re-scaling achieves the best results with Sil/Manhattan and Sil/Minkowski for $p$ between 1.4 and 2, with optimal results (out of all methods) between 1.5 and 1.7.  iMWK-Means with re-scaling followed by K-Means with Sil/Manhattan beats K-MSM for $p\ge 1.4$. Regarding the percentage of finding the true number of clusters (Figure \ref{fig:perc}), K-MSM shows excellent results that were not achieved by any index together with iMWK-Means without and with re-scaling, both of which rank also below K-MSE. 
However, the best method is iMWK-Means with re-scaling followed by K-Means for $p>2$ together with Sil/Manhattan. With several other indexes, iMWK-Means with re-scaling followed by K-Means for a wide range of values of $p$ ranks ahead of plain K-Means with any index. Regarding the relative error of the number of clusters, results are mostly in between what is showed in the two Figures.

\subsection{Other scenarios}\label{ssimother}
The standard scenarios in Section \ref{ssimstan} are based on spherical Gaussian clusters, uniform noise and uniform cluster proportions. Because there is a virtually unlimited variety of potentially interesting and relevant scenarios, it is necessary to restrict even a comprehensive simulation study in one way or another. Just in order to widen our scope, we ran two limited experiments exploring some other aspects. In both of these experiments we restricted ourselves to the situation with 12 relevant features, 6 noise features and 3 clusters. The three clusters here have proportions of 0.5, 0.3 and 0.2. In the first experiment we used t-distributions with 3 degrees of freedom for both cluster and noise in order to investigate the effect of outliers. In the second experiment we used the Gaussian distribution for both relevant features and noise, but we introduced correlations of $\sigma^2/3$ between the relevant features. All other parameters were chosen as in Section \ref{ssimstan} (using $\sigma^2$ as scale parameters instead of variances for the t-distributions). Again we ran 50 replicates in each experiment. We focus on results with the adjusted Rand index.

In the first experiment, the best baseline method is once more K-MSM with average Rand index of 0.799. iMWK-Means imroves on this with Hartigan and $p=1.7, 1.8$ and 2. The best value with $p=1.8$ is 0.823. iMWK with re-scaling gives results better than K-Means with any index for $p$ between 1.7 and 2 with all three versions of Sil, but none of them beats K-MSM. Finally, the best result (0.828) is achieved by iMWK-Means with re-scaling followed by K-Means, Sil/Manhattan and $p=2$ (note that this combination is also the best one regarding the percentage of estimating the true number of clusters); actually results of this combination are better than K-MSM for $p\ge 1.5$.  

In the second experiment, the best baseline method is K-MSE (adjusted Rand index 0.865), whereas K-MSM achieves 0.847. The best result overall (0.868) is achieved by iMWK-Means with re-scaling followed by K-Means with Sil/Minkowski and $p=3$. K-MSE was generally hard to beat here; results of 0.866 are achieved by iMWK-Means and iMWK-Means with re-scaling followed by K-Means, both with Sil/Euclidean and Sil/Minkowski and $p=2.5$; most results of the three versions of iMWK-Means together with the three versions of Sil and $p\ge 1.4$ are better than K-MSM here. The best overall result regarding the percentage of estimating the true number of clusters is achieved by iMWK-Means with re-scaling followed by K-Means with Sil/Manhattan and $p=3$, which is somewhat surprising, because this combination did not perform particularly good regarding the adjusted Rand index.

\subsection{Computation times} 
Some computation times are given in Table \ref{Tab:Time}. These experiments were run in one of the cores of an Intel i7-3630QM in a 64 bits computer with 8 GB of RAM running Ubuntu 14.04 and MATLAB R2013.
At $p=2$, iMWK-Means is always considerably faster than at $p=1.4$, because in the former case each centroid is found using the component-wise mean of the entities in its cluster. This is considerably faster than using our steepest descent method. In general, although computation times for our method are substantially higher than for K-Means, the computational burden is still rather low in absolute terms. 

\begin{table}[t]
\caption{Average time, in seconds, of 100 runs under each of the configurations.}
\begin{center}
\tabcolsep=0.04cm 
\begin{tabular}{@{\extracolsep{2pt}}lccccccc@{}}
&K-Means&\multicolumn{4}{c}{iMWK-Means}\\
&&\multicolumn{2}{c}{No Re-scaling}&\multicolumn{2}{c}{With Re-scaling}\\
\cline{3-4}
\cline{5-6}
&&p=1.4&p=2.0&p=1.4&p=2.0\\
1000x8-2&0.0048&0.0944&0.0100&0.0949&0.0101\\
1000x8-2 + 8NF&0.0055&0.2031&0.0237&0.2034&0.0230\\
1000x12-3&0.0133&0.4378&0.0290&0.4405&0.0288\\
1000x12-3 + 12NF&0.0119&0.5737&0.0424&0.5754&0.0405\\
1000x16-4&0.0096&0.3479&0.0363&0.3500&0.0357\\
1000x16-4 + 16NF&0.0524&1.1590&0.1786&1.1474&0.1706\\
1000x20-5&0.0139&0.5392&0.0324&0.5301&0.0309\\
1000x20-5 + 20NF&0.0241&1.4572&0.1352&1.4356&0.1328\\
\end{tabular}
\end{center}
\label{Tab:Time}
\end{table}

\section{Conclusion}

In this paper we have introduced three feature re-scaling based methods with the view of improving the estimation of the number of clusters and generally clustering with unknown number of cluster by using them together with a number of popular cluster validity indexes. The methods are based on the feature weights obtained by the intelligent Minkowski Weighted K-Means. The re-scaling can be implicit thanks to the use of the weighted Minkowski distance in (\ref{Eq:AdjustedDistanceMeasure}), or explicit, in which case the value of a feature in a given entity is multiplied by a re-scaling factor. This may be followed by further clustering or only used for computation of the cluster validity indexes. The re-scaling takes into account to which cluster each entity is assigned, as well as each feature and its relative dispersion to the other features in the data set.

We have assessed our methods by presenting experiments on 700 synthetically generated data sets, with and without noise features. In the presence of noise features, our methods improve the performance of most cluster validity indexes such as the Silhouette (using the squared Euclidean, Manhattan, and Minkowski distances), Dunn's (using the Euclidean and Minkowski distances), and Calinski-Harabasz indexes for suitable values of Minkowski's $p$; $p$ between 1.4 and 1.8 led to good results, but $p\ge 2$ and particularly $p=3$ together with Silhouette (Manhattan) was a strong choice as well. 

Out of the methods proposed here, the plain iMWK-Means algorithm performed best without noise. With noise, iMWK-Means with explicit feature re-scaling followed by K-Means was superior regarding estimating the number of clusters, although iMWK-Means with explicit feature re-scaling without K-Means achieved better results regarding the adjusted Rand-index, which means that it seemed to go for a number of clusters different from the true one in cases in which the clustering with the true number may have produced some misclassifications, whereas a clustering with a different number of clusters could adapt better to the true clustering structure.

Future research will address the estimation of $p$ in different scenarios, and methods to initialize K-Means, avoiding having to run it a number of times as we did in our experiments. The re-scaling approach could also benefit other clustering methods such as Partitioning around Medoids \citep{Kaufman90}.
%
\bibliography{References0315}

\begin{thebibliography}{10}
\expandafter\ifx\csname url\endcsname\relax
  \def\url#1{\texttt{#1}}\fi
\expandafter\ifx\csname urlprefix\endcsname\relax\def\urlprefix{URL }\fi

\bibitem{arbelaitz2012extensive}
O.~Arbelaitz, I.~Gurrutxaga, J.~Muguerza, J.~M. P{\'e}rez, I.~Perona, An
  extensive comparative study of cluster validity indices, Pattern Recognition
  46~(1) (2012) 243–--256.

\bibitem{ball1967clustering}
G.~H. Ball, D.~J. Hall, A clustering technique for summarizing multivariate
  data, Behavioral Science 12~(2) (1967) 153--155.

\bibitem{bezdek1998some}
J.~C. Bezdek, N.~R. Pal, Some new indexes of cluster validity, Systems, Man,
  and Cybernetics, Part B: Cybernetics, IEEE Transactions on 28~(3) (1998)
  301--315.

\bibitem{calinski1974dendrite}
T.~Cali{\'n}ski, J.~Harabasz, A dendrite method for cluster analysis,
  Communications in Statistics-theory and Methods 3~(1) (1974) 1--27.

\bibitem{chan2004optimization}
E.~Y. Chan, W.~K. Ching, M.~K. Ng, J.~Z. Huang, An optimization algorithm for
  clustering using weighted dissimilarity measures, Pattern recognition 37~(5)
  (2004) 943--952.

\bibitem{chiang2007experiments}
M.~M.-T. Chiang, B.~Mirkin, Experiments for the number of clusters in k-means,
  in: Progress in Artificial Intelligence, vol. 4874 of Lecture Notes in
  Computer Science, Springer, 2007, pp. 395--405.

\bibitem{chiang2010intelligent}
M.~M.-T. Chiang, B.~Mirkin, Intelligent choice of the number of clusters in
  k-means clustering: an experimental study with different cluster spreads,
  Journal of classification 27~(1) (2010) 3--40.

\bibitem{arthur2006worst}
A.~David, S.~Vassilvitskii, Worst-case and smoothed analysis of the icp
  algorithm, with an application to the k-means method, in: The 47th Annual
  IEEE Symposium on Foundations of Computer Science, IEEE, California, USA,
  2006, pp. 153--164.

\bibitem{de2013empirical}
R.~C. de~Amorim, An empirical evaluation of different initializations on the
  number of k-means iterations, in: Advances in Artificial Intelligence,
  Lecture Notes in Computer Science, Springer, 2013, pp. 15--26.

\bibitem{de2012initializations}
R.~C. de~Amorim, P.~Komisarczuk, On initializations for the minkowski weighted
  k-means, in: Advances in Intelligent Data Analysis XI, vol. 7619 of Lecture
  Notes in Computer Science, Springer, 2012, pp. 45--55.

\bibitem{cordeiro2011minkowski}
R.~C. de~Amorim, B.~Mirkin, Minkowski metric, feature weighting and anomalous
  cluster initializing in k-means clustering, Pattern Recognition 45~(3) (2012)
  1061--1075.

\bibitem{dudoit2002prediction}
S.~Dudoit, J.~Fridlyand, A prediction-based resampling method for estimating
  the number of clusters in a dataset, Genome biology 3~(7) (2002) 1--21.

\bibitem{dunn1973fuzzy}
J.~C. Dunn, A fuzzy relative of the isodata process and its use in detecting
  compact well-separated clusters, Journal of Cybernetics 3~(3) (1973) 32--57.

\bibitem{gasch2002exploring}
A.~P. Gasch, M.~B. Eisen, Exploring the conditional coregulation of yeast gene
  expression through fuzzy k-means clustering, Genome Biol 3~(11) (2002) 1--22.

\bibitem{halkidi2001clustering}
M.~Halkidi, Y.~Batistakis, M.~Vazirgiannis, On clustering validation
  techniques, Journal of Intelligent Information Systems 17~(2-3) (2001)
  107--145.

\bibitem{hartigan1975clustering}
J.~A. Hartigan, Clustering algorithms, John Wiley \& Sons, Inc., 1975.

\bibitem{Hennig_Liao_2013}
C.~Hennig, T.~F. Liao, How to find an appropriate clustering for mixed type
  variables with application to socioeconomic stratification (with discussion),
  Journal of the Royal Statistical Science, Series C (Applied Statistics) 62
  (2013) 309--369.

\bibitem{huang2005automated}
J.~Z. Huang, M.~K. Ng, H.~Rong, Z.~Li, Automated variable weighting in k-means
  type clustering, Pattern Analysis and Machine Intelligence, IEEE Transactions
  on 27~(5) (2005) 657--668.

\bibitem{huang2008weighting}
J.~Z. Huang, J.~Xu, M.~Ng, Y.~Ye, Weighting method for feature selection in
  k-means, in: H.~Liu, H.~Motoda (eds.), Computational Methods of Feature
  Selection, Data Mining \& Knowledge Discovery, Chapman \& Hall/CRC, 2008, pp.
  193--209.

\bibitem{hubertarabie1985rand}
L.~Hubert, P.~Arabie, Comparing partitions, Journal of Classification 2~(2)
  (1985) 193--218.

\bibitem{jain2010data}
A.~K. Jain, Data clustering: 50 years beyond k-means, Pattern Recognition
  Letters 31~(8) (2010) 651--666.

\bibitem{jain1988algorithms}
A.~K. Jain, R.~C. Dubes, Algorithms for clustering data, Prentice-Hall, Inc.,
  1988.

\bibitem{Kaufman90}
L.~Kaufman, P.~Rousseeuw, Finding Groups in Data, Wiley, New York, 1990.

\bibitem{macqueen1967some}
J.~MacQueen, Some methods for classification and analysis of multivariate
  observations, in: Proceedings of the fifth Berkeley symposium on mathematical
  statistics and probability, vol.~1, California, USA, 1967, pp. 281--297.

\bibitem{milligan1985examination}
G.~W. Milligan, M.~C. Cooper, An examination of procedures for determining the
  number of clusters in a data set, Psychometrika 50~(2) (1985) 159--179.

\bibitem{mirkin2012clustering}
B.~Mirkin, Clustering: A Data Recovery Approach, Computer Science and Data
  Analysis, CRC Press, London, UK, 2012.

\bibitem{pollard2002method}
K.~S. Pollard, M.~J. Van Der~Laan, A method to identify significant clusters in
  gene expression data, in: Proceedings of the 6th World Multiconference on
  Systemics, Cybernetics and Informatics, bepress, Orlando, USA, 2002, pp.
  318--325.

\bibitem{rousseeuw1987silhouettes}
P.~J. Rousseeuw, Silhouettes: a graphical aid to the interpretation and
  validation of cluster analysis, Journal of computational and applied
  mathematics 20 (1987) 53--65.

\bibitem{steinley2006k}
D.~Steinley, K-means clustering: A half-century synthesis, British Journal of
  Mathematical and Statistical Psychology 59~(1) (2006) 1--34.

\bibitem{steinley2007initializing}
D.~Steinley, M.~J. Brusco, Initializing k-means batch clustering: a critical
  evaluation of several techniques, Journal of Classification 24~(1) (2007)
  99--121.

\bibitem{sturn2002genesis}
A.~Sturn, J.~Quackenbush, Z.~Trajanoski, Genesis: cluster analysis of
  microarray data, Bioinformatics 18~(1) (2002) 207--208.

\bibitem{vedaldi2010vlfeat}
A.~Vedaldi, B.~Fulkerson, Vlfeat: An open and portable library of computer
  vision algorithms, in: A.~Del~Bimbo, S.~Chang, A.~Smeulders (eds.),
  Proceedings of the international conference on Multimedia, ACM, Firenze,
  Italy, 2010, pp. 1469--1472.

\end{thebibliography}
\end{document}